\documentclass[runningheads]{llncs}
\usepackage{graphicx}
\usepackage{amsmath,amssymb,amsopn,amstext,amsfonts}
\usepackage{color}
\usepackage{pdfsync}
\usepackage{balance}
\usepackage{color}
\usepackage{algorithm}
\usepackage{multirow}
\usepackage{algorithmic}
\usepackage{epstopdf}
\usepackage{xcolor}
\usepackage{mathtools}
\usepackage[]{algorithmic}
\usepackage[]{algorithm}
\usepackage{multirow}
\usepackage{subfig}
\usepackage{float}
\usepackage{diagbox}
\usepackage{makecell}
\usepackage{caption}

\begin{document}

\title{Modeling Varying Camera-IMU Time Offset in Optimization-Based Visual-Inertial Odometry}
\titlerunning{Modeling Varying Camera-IMU Time Offset in Optimization-Based VIO}
\author{Yonggen Ling \and Linchao Bao \and Zequn Jie \and Fengming Zhu \and Ziyang Li \and \\Shanmin Tang \and Yongsheng Liu \and Wei Liu \and Tong Zhang}
\authorrunning{Ling \textit{et al.}}
\institute{Tencent AI Lab, China
	\email{ylingaa@connect.ust.hk,\{linchaobao,zequn.nus,fridazhu\}@gmail.com, \{tzeyangli,mickeytang,kakarliu\}@tencent.com, wl2223@columbia.edu,
tongzhang@tongzhang-ml.org}}

\maketitle

\begin{abstract}
Combining cameras and inertial measurement units (IMUs) has been proven effective in motion tracking, as these two sensing modalities offer complementary characteristics that are suitable for fusion. While most works focus on global-shutter cameras and synchronized sensor measurements, consumer-grade devices are mostly equipped with rolling-shutter cameras and suffer from imperfect sensor synchronization. In this work, we propose a nonlinear optimization-based monocular visual inertial odometry (VIO) with varying camera-IMU time offset modeled as an unknown variable. Our approach is able to handle the rolling-shutter effects and imperfect sensor synchronization in a unified way. Additionally, we introduce an efficient algorithm based on dynamic programming and red-black tree to speed up IMU integration over variable-length time intervals during the optimization. An uncertainty-aware initialization is also presented to launch the VIO robustly. Comparisons with state-of-the-art methods on the Euroc dataset and mobile phone data are shown to validate the effectiveness of our approach.

\end{abstract}

\section{Introduction}
\label{sec:introduction}
Online, robust, and accurate localization is the foremost important component for many applications, such as autonomous navigation of mobile robots, online augmented reality, and real-time localization-based service.
A monocular VIO that consists of one IMU and one camera is particularly suitable for this task as these two sensors are cheap, ubiquitous, and complementary.
However, a VIO works only if both visual and inertial measurements are aligned spatially and temporally.
This requires that both sensor measurements are synchronized and sensor extrinsics between sensors are known.
While online sensor extrinsic calibration has gained lots of discussions in recent works, VIO with inperfect synchronization is less explored.
Historically, some works \cite{jacovitti93,kelly2010} calibrate the sensor time offsets offline, and assume that these parameters are not changed in next runs.
In real cases, time offsets change over time due to the variation of the system processing payload and sensor jitter.
Other works \cite{limingyangicra13,Guo-RSS-14,ijrr_mingyang14} calibrate sensor time offsets online in an extended Kalman filter (EKF) framework.
However, these methods suffer from the inherent drawbacks of filtering based approaches.
They require a good prior about the initial system state (such as poses, velocities, biases, the camera-IMU extrinsic/time offsets) such that the estimation at each update step converges close to the global minimum.
In contrast to filtering based approaches, nonlinear optimization-based methods \cite{jung_taylor,leuRSS2013,yang2016vins,DonMou1210} iteratively re-linearize all nonlinear error costs from visual and inertial constraints to better treat the underlying nonlinearity, leading to increased tracking robustness and accuracy.
However, introducing time offsets in a nonlinear optimization framework is non-trivial since the visual constraints are varying as they depend on the estimated time offsets that are varied between iterations.
Another critical problem of VIO is the use of rolling-shutter cameras. Unlike global-shutter cameras that capture all rows of pixels at one time instant, rolling-shutter cameras capture each row of pixels at a different time instant.
The rolling-shutter effect on captured images causes a significant geometry distortion if the system movement is fast.
Without taking the rolling-shutter effect into account, the estimation performance degrades rapidly.
Unfortunately, most consumer-grade cameras (such as cameras on mobile phones) are rolling-shutter cameras.
If we optimize camera poses at every readout time of rolling-shutter cameras, the computational complexity will be intractable.

\begin{figure}[!t]	
	\centering	
	\includegraphics[width=0.8\textwidth]{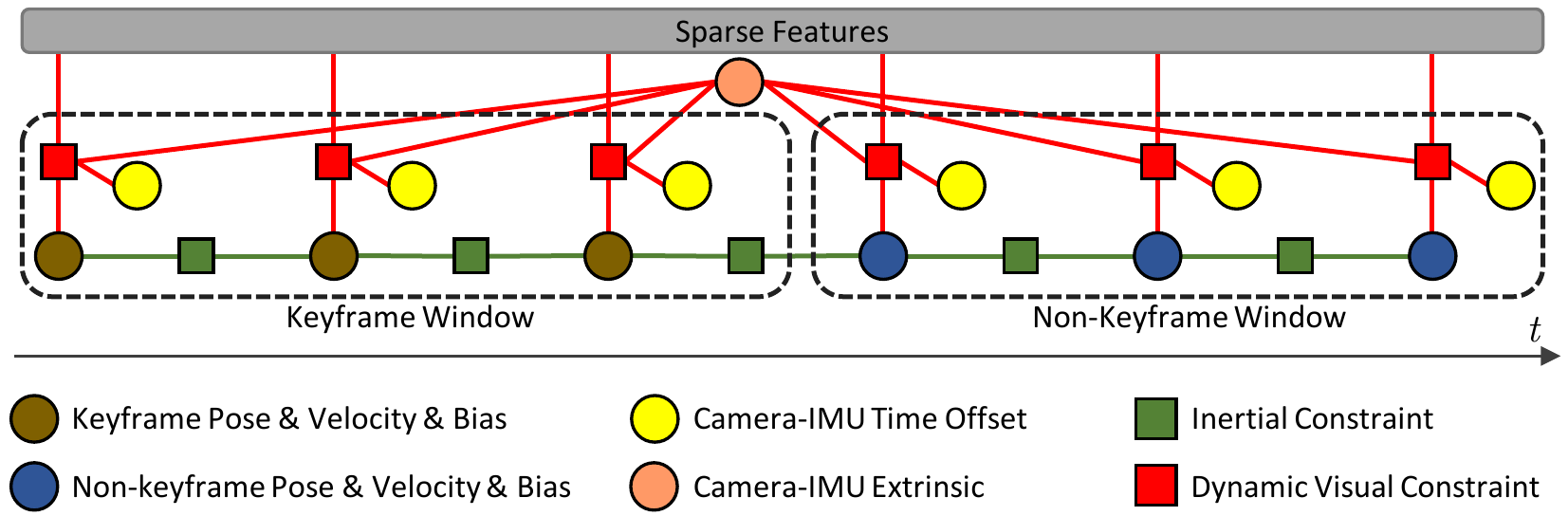}
	\caption{The graph representation of our model. All variables (circles) and constraints (squares) in both the keyframe window and non-keyframe window are involved in the optimization. Note that modeling the camera-IMU time offset for each frame raises computational difficulties during the optimization, since the computation of visual constraints depends on the estimated time offset. In other words, the visual constraints in our model are ``dynamic'' due to the varying camera-IMU time offset.}
	\label{fig:graph}
\end{figure}

To the best of our knowledge, we are the first to propose a nonlinear optimiza\-tion-based VIO to overcome the difficulties mentioned above.
The graph representation of our model is shown in Fig. \ref{fig:graph}.
Different from prior VIO algorithms based on nonlinear optimization,
we incorporate an unknown, dynamically changing time offset for each camera image (shown as yellow circle in Fig. \ref{fig:graph}).
The time offset is jointly optimized together with other variables like poses, velocities, biases, and camera-IMU extrinsics.
We show that by modeling the time offset as a time-varying variable,
imperfect camera-IMU synchronization and rolling-shutter effects can be handled in a unified formulation (Sect. \ref{subsec:rollingshuttercomp} and Sect. \ref{subsec:offset_model}).
We derive the Jacobians involved in the optimization after introducing the new variable (Sect. \ref{subsec:modelopt}),
and show that the varying time offset brings computational challenges of pose computation over variable-length time intervals.
An efficient algorithm based on dynamic programming and red-black tree is proposed to ease these difficulties (Sect. \ref{subsec:efficientalgo}).
Finally, since the nonlinear optimization is based on linearization, an initial guess is required for optimization bootstrap. A poor initialization may lead to a decrease of VIO robustness and accuracy. To improve the robustness of the system bootstrap, we present an initialization scheme, which takes the uncertainty of sensor measurements into account and better models the underlying sensor noises. Main contributions of this paper are as follows:

\begin{itemize}
 		\item We propose a nonlinear optimization-based VIO with varying camera-IMU time offset modeled as an unknown variable, to deal with the rolling-shutter effects and online temporal camera-IMU calibration in a unified framework.
 		\item We design an efficient algorithm based on dynamic programming and red-black tree to speed up the IMU integration over variable-length time intervals, which is needed during optimization.
 		\item We introduce an uncertainty-aware initialization scheme to improve the robustness of the VIO bootstrap.
 	\end{itemize}
Qualitative and quantitative results on the Euroc dataset with simulated camera-IMU time offsets and real-world mobile phone data are shown to demonstrate the effectiveness of the proposed method.

\section{Related Work}
\label{sec:related_work}

The idea of VIO goes back at least to the work \cite{roumeliotis02icra} proposed by Roumeliotis \textit{et al.}  based on filtering and the work \cite{jung_taylor} proposed by Jung and Taylor based on nonlinear optimization. Subsequently, lots of work have been published based on an exemplar implementation of filtering based approaches, called EKF \cite{SheMulMic1305,LiMou1305,HesKotBow1402,msckf}. EKFs predict latest motions using IMU measurements and perform updates according to the reprojection errors from visual measurements. To bound the algorithmic complexity, many works follow a loosely coupled fashion \cite{WeiAchLyn1205,simon13,SheMulMic1305,LingShen15,LingShen16,auro17,wang18}. Relative poses are firstly estimated by IMU propagations as well as vision-only structure from motion algorithms separately. They are then fused together for motion tracking.
Alternatively, approaches in a tightly coupled fashion optimize for estimator states using raw measurements from IMUs and cameras \cite{msckf,LiMou1305}. They take the relations among internal states of different sensors into account, thus achieving higher estimation accuracy than loosely coupled methods at the cost of a higher computational complexity.
Additionally, to benefit from increased accuracy offered by relinearization, nonlinear optimization-based methods iteratively minimize errors from both inertial measurements and visual measurements \cite{leuRSS2013,yang2016vins,DonMou1210}. The main drawback of nonlinear optimization-based methods is their high computational complexity due to repeated linearizations, which can be lessened by limiting the variables to optimize and utilizing structural sparsity of the visual-inertial problem \cite{leuRSS2013}.

Recent approaches on VIOs consider the problem of spatial or temporal camera-IMU calibration. The camera-IMU relative transformation is calibrated offline \cite{paul_joem_13} using batch optimization, or online by including it into the system state for optimization \cite{LiMou1305,WeiAchLyn1205,yangShen15}. The temporal calibration between the camera and the IMU is a less-explored topic \cite{jacovitti93,kelly2010,limingyangicra13,Guo-RSS-14}. Jacovitti \textit{et al.} \cite{jacovitti93} estimate the time-offset by searching the peak that maximizes the correlation of different sensor measurements. Kelly \textit{et al.} \cite{kelly2010} firstly estimated rotations from different sensors independently, and then temporally aligned these rotations in the rotation space. However, both \cite{jacovitti93,kelly2010} cannot estimate time-varying time offsets. Li \textit{et al.} \cite{limingyangicra13} adopted a different approach. They assume constant velocities around local trajectories. Time offsets are included in the estimator state vector, and optimized together with other state variables within an EKF framework. Instead of explicitly optimizing the time offset, Guo \textit{et al.} \cite{Guo-RSS-14} proposed an interpolation model to account for the pose displacements caused by time offsets.

While most works on VIOs use global-shutter cameras, deployments on consumer devices drive the need for using rolling-shutter cameras. A straightforward way to deal with rolling-shutter effects is to rectify images as if they are captured by global-shutter cameras, such as the work \cite{KleMur09} proposed by Klein and Murray that assumes a constant velocity model and corrects distorted image measurements in an independent thread. For more accurate modeling, some approaches extend the camera projection function to take rolling-shutter effects into account. They represent local trajectories using zero order parameterizations \cite{limingyangicra13,limingyangicra13_RS} or higher order parameterizations \cite{splinefusion}. Instead of modeling the trajectories, \cite{LiMou14rollingshutterIJRR} predicts the trajectories using IMU propagation and models the prediction errors of the estimated trajectories. These errors are represented as a weighted sum of temporal basis functions.

\section{Preliminaries}\label{sec:preliminaries}
In this section we briefly review the preliminaries of the nonlinear optimization framework used in our model. For detailed derivations, please refer to \cite{SheMicKum1505,leuRSS2013}.

We begin by giving notations. We consider $(\cdot)^w$ as the earth's inertial frame, and $(\cdot)^{b_k}$ and $(\cdot)^{c_k}$ as the IMU body frame and camera frame while taking the $k^{th}$ image, respectively.
We use $\mathbf{p}^X_Y$, $\mathbf{v}^X_Y$, and $\mathbf{R}^X_Y$ to denote the 3D position, velocity, and rotation of frame $Y$ w.r.t frame $X$, respectively.
The corresponding quaternion ($\mathbf{q}^X_Y=\left [q_x,\,q_y,\,q_z,\,q_w \right ]$) for rotation is in Hamilton notation in our formulation.
We assume that the intrinsic of the monocular camera is calibrated beforehand with known focal length and principle point.
The relative translation and rotation between the monocular camera and the IMU are $\mathbf{p}_b^c$ and $\mathbf{q}_b^c$.
The system-recorded time instant for taking the $k^{th}$ image is $t_k$, while the image is actually captured at $\tilde{t}_k = t_k + \Delta t^o_{k}$,
with an unknown time offset $\Delta t^o_{k}$ due to inaccurate timestamps.
Note that the time offset $\Delta t^o_{k}$ is generally treated as a known constant in other optimization-based VIO algorithms, whereas it is modeled as an unknown variable for each image in our model.

In a sliding-window nonlinear optimization framework, the full state is usually encoded as
$\mathcal{X} = [ \mathbf{x}_{b_0} \ ... \ \mathbf{x}_{b_k} \  ... \ \mathbf{x}_{b_n} \  \mathbf{f}_0^{w} \ ... \ \mathbf{f}_j^{w} \ ... \ \mathbf{f}_m^{w} \ \mathbf{p}_{c}^b \  \mathbf{q}_{c}^b ]$, where a sub-state $\mathbf{x}_{b_k} = [\mathbf{p}_{b_k}^w, \mathbf{v}_{b_k}^w, \mathbf{q}_{b_k}^w, \mathbf{b}_a^{b_k}, \mathbf{b}_\omega^{b_k}]$ consists of the position, velocity, rotation, linear acceleration bias, and angular velocity bias at $t_k $,
$\mathbf{f}_j^{w}$ is the 3D Euclidean position of feature $j$ in the world coordinate, and $\mathbf{p}_{c}^b$ as well as $\mathbf{q}_{c}^b $ are the camera-IMU extrinsics.
Finding the MAP estimate of state parameters is equivalent to minimizing the sum of the Mahalanobis norm of all measurement errors:
\begin{flalign}
	\label{equ:MAP}
	\underset{\mathcal{X}} {\min}  \ \  || \mathbf{b}_p - \mathbf{H}_p \mathcal{X}||^2 + \sum_{\hat{\mathbf{z}}_{k+1}^{k} \in S_i} || r_i(\hat{\mathbf{z}}_{k+1}^{k}, \mathcal{X} ) ||_{\mathbf{\Sigma}_{k+1}^{k} }^2 + \sum_{\hat{\mathbf{z}}_{ik} \in S_c} || r_c (\hat{\mathbf{z}}_{ik}, \mathcal{X}) ||_{\mathbf{\Sigma}_{c} }^2,
\end{flalign}
where $\mathbf{b}_p$ and $\mathbf{H}_p$ are priors obtained via marginalization \cite{leuRSS2013},
$S_i$ and $S_c$ are the sets of IMU and camera measurements,
with the corresponding inertial and visual constraints modeled by residual functions
$r_i(\hat{\mathbf{z}}_{k+1}^{k}, \mathcal{X} )$ and $r_c (\hat{\mathbf{z}}_{ik}, \mathcal{X})$, respectively.
The corresponding covariance matrices are denoted as $\mathbf{\Sigma}_{k+1}^{k}$ and $\mathbf{\Sigma}_{c}$.

To derive the inertial residual term $r_i(\hat{\mathbf{z}}_{k+1}^{k}, \mathcal{X} )$ in Eq. \eqref{equ:MAP}, the IMU propagation model needs to be derived from the kinematics equation first, that is
\begin{flalign}
\label{equ:imu_propagation}
\mathbf{p}_{b_{k+1}}^{w} &= \mathbf{p}_{b_k}^{w} + \mathbf{v}^w_{b_k} \Delta t_k - \frac{1}{2}\mathbf{g}^{w}  \Delta t_k ^2 + \mathbf{R}_{b_k}^{w}  \hat{\boldsymbol{\alpha}}_{k+1}^{k}, \nonumber \\
\mathbf{v}^w_{b_{k+1}} &= \mathbf{v}^w_{b_{k}}   - \mathbf{g}^{w} \Delta t_k  + \mathbf{R}_{b_k}^{w} \hat{\boldsymbol{\beta}}_{k+1}^{k},\\
\mathbf{q}_{k+1}^{w} &= \mathbf{q}_{k}^{w} \otimes \hat{\mathbf{q}}_{k+1}^{k}, \nonumber
\end{flalign}
where $\Delta t_k = t_{k+1} - t_{k}$, $\mathbf{g}^{w} = {[0, 0, 9.8]}^T$ is the gravity vector in the earth's inertial frame, and $\hat{\mathbf{z}}_{k+1}^{k} = \{\hat{\boldsymbol{\alpha}}_{k+1}^{k}, \hat{\boldsymbol{\beta}}_{k+1}^{k}, \hat{\mathbf{q}}_{k+1}^{k} \}$ as well as its covariance $\mathbf{\Sigma}_{k+1}^{k}$ can be obtained by integrating linear accelerations $\mathbf{a}^{b_t}$ and angular velocities $ \boldsymbol{\omega}^{b_t}$ \cite{forster15}.
Then the inertial residual term can be derived as:
\begin{flalign}
r_i( \hat{\mathbf{z}}_{k+1}^{k} , \mathcal{X} ) =
\begin{bmatrix}
\mathbf{R}_{w}^{b_k} ( \mathbf{p}_{b_{k+1}}^w - \mathbf{p}_{b_k}^w - \mathbf{v}_{b_k}^{w} \Delta t_k + \frac{1}{2} \mathbf{g}^{w}\Delta t_k ^2  ) - \hat{\boldsymbol{\alpha}}_{{k+1}}^{k} \\
\mathbf{R}_{w}^{b_k} (  \mathbf{v}_{b_{k+1}}^w  -  \mathbf{v}_{b_{k}}^w  + \mathbf{g}^{w} \Delta t_k )  -  \hat{\boldsymbol{\beta}}_{k+1}^{k} \\
(\hat{\mathbf{q}}_{{k+1}}^{{k}})^{-1}  (\mathbf{q}_{{ b_k}}^{{w}})^{-1}  \mathbf{q}_{b_{k+1}}^{{w}}
\end{bmatrix}.
\label{equ:inertial_residual}
\end{flalign}

The visual residual term $r_c (\hat{\mathbf{z}}_{ik}, \mathcal{X})$ in Eq. \eqref{equ:MAP} is defined by the projection errors of tracked sparse features, 
which can be obtained using ST corner detector \cite{ShiTom9406} and tracked across sequential images using sparse optical flow \cite{baker2004lucas}.
Note that, to handle the rolling-shutter effect, the generalized epipolar geometry \cite{generalized_epipolar_geometry} can be adopted as the fitted model in the RANSAC outlier removal procedure during correspondences establishment.
Suppose a feature $\mathbf{f}_i^w$ in the world coordinate, following the pinhole model, its projection $\mathbf{u}_i^k$ on the $k$ frame is:
{\begin{flalign}
\label{equ:globalvisualproj}
\mathbf{u}_i^k =  \begin{bmatrix}
x_i^{c_k} / z_i^{c_k} \\
y_i^{c_k}  / z_i^{c_k}\\
\end{bmatrix},  \ \text{where} \   \mathbf{f}_i^{c_k} =   \begin{bmatrix}
x_i^{c_k} \\
y_i^{c_k} \\
z_i^{c_k} \\
\end{bmatrix}
= \mathbf{R}^{c}_{b} ( \mathbf{R}_w^{b_k} (\mathbf{f}_i^w - \mathbf{p}_{b_k}^w) - \mathbf{p}_c^{b} ).
\end{flalign}
Then the projection error is $ r_c (\hat{\mathbf{z}}_{ik}, \mathcal{X}) = \mathbf{u}_i^k - \hat{\mathbf{u}}_i^k $, where $ \hat{\mathbf{u}}_i^k$ is the tracked feature location.
The covariance matrix $\mathbf{\Sigma}_{c}$ is set according to the tracking accuracy of the feature tracker.
By linearizing the cost function in Eq. \eqref{equ:MAP} at the current best estimation $\hat{\mathcal{X}}$ with respect to error state $\mathcal{\delta X}$,
the nonlinear optimization problem is solved via iteratively minimizing the following linear system over $\mathcal{\delta X}$
and updating the state estimation $\hat{\mathcal{X}} $ by $\hat{\mathcal{X}} \leftarrow \hat{\mathcal{X}} + \mathcal{\delta X} $ until convergence:
\begin{flalign}
		\label{equ:MAP_error}
		\underset{\mathcal{\delta X}} {\min}  \  || \mathbf{b}_p - \mathbf{H}_p ( \hat{\mathcal{X}} + \delta\mathcal{X}) ||^2  & + \sum_{\hat{\mathbf{z}}_{k+1}^{k} \in S_i}  || r_i(\hat{\mathbf{z}}_{k+1}^{k}, \hat{\mathcal{X}} ) + \mathbf{H}_{k} \mathcal{\delta X} ||_{\mathbf{\Sigma}_{k+1}^{k} }^2  \nonumber\\
		& + \sum_{\hat{\mathbf{z}}_{ik} \in S_c} || r_c (\hat{\mathbf{z}}_{ik}, \hat{\mathcal{X}}) + \mathbf{H}_{k}^i \mathcal{\delta X} ||_{\mathbf{\Sigma}_{c} }^2,
\end{flalign}
where $\mathbf{H}_{k}$ and $ \mathbf{H}_{k}^i$ are Jacobians of the inertial and visual residual functions.

\section{Modeling Varying Camera-IMU Time Offset}\label{sec:proposedmethod}

In this section, we first show that rolling-shutter effects can be approximately compensated by modeling a camera-IMU time offset (Sect. \ref{subsec:rollingshuttercomp}).
Then we present our time-varying model for the offset (Sect. \ref{subsec:offset_model}) and the derivation of the Jacobian for optimization after introducing the time offset variable (Sect. \ref{subsec:modelopt}).
Finally, an efficient algorithm is described to accelerate the IMU integration over variable-length time intervals (Sect. \ref{subsec:efficientalgo}), required by the optimization.

\subsection{Approximate Compensation for Rolling-Shutter Effects}\label{subsec:rollingshuttercomp}
\begin{figure}[!t]
	\centering
	\includegraphics[width=0.9\textwidth]{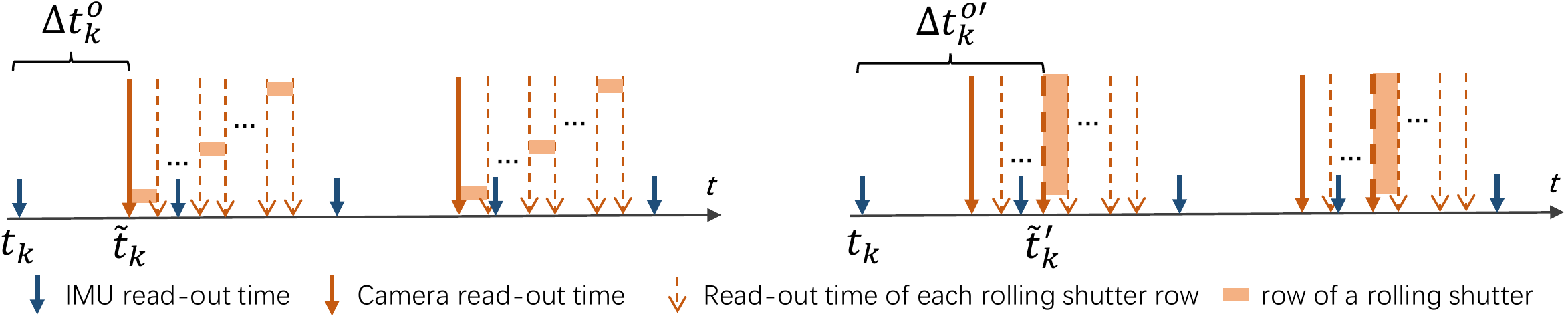} \ \ \ \ \
	\includegraphics[width=0.55\textwidth]{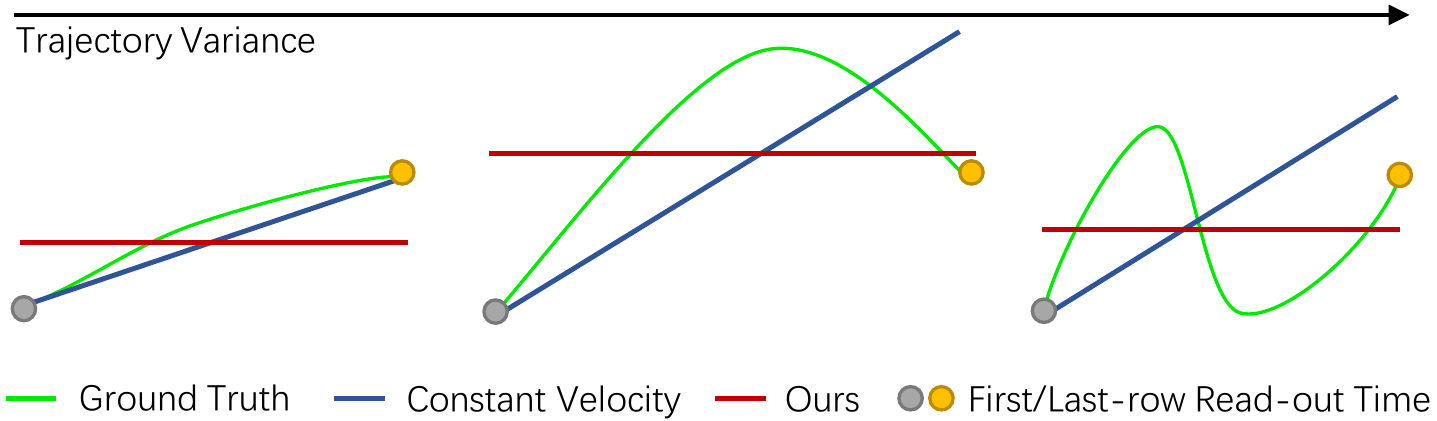}
	\caption{The illustration of a camera-IMU sensor suite with a rolling-shutter camera and imperfect camera-IMU synchronization. We `average' the rolling shutter readout time instants and approximate the rolling-shutter images (top-left) with global-shutter images captured at the `middle-row' readout time of the rolling-shutter cameras (top-right). The position of this `middle-row' is optimized to be the expected position of the local ground truth trajectory (bottom). }
	\label{fig:rolling_shutter}
\end{figure}

Fig.~\ref{fig:rolling_shutter}.(a) shows a camera-IMU suite with a rolling-shutter camera and imperfect camera-IMU synchronization.
The system-recorded time instant for taking the $k^{th}$ image is $t_k$,
which serves as our time reference for the retrieval of IMU data.
Due to imperfect sensor synchronization (or inaccurate timestamps),
the image is actually captured at $\tilde{t}_k = t_k + \Delta t^o_{k}$,
with an unknown time offset $\Delta t^o_{k}$.
With a rolling-shutter camera, this means that $\tilde{t}_k$ is the time instant when the camera starts to read out pixels row by row.
Instead of modeling local trajectories using constant velocities \cite{Guo-RSS-14,KleMur09,limingyangicra13_RS}, we model them with constant poses, which are expected poses of local trajectories (Fig.~\ref{fig:rolling_shutter}).
With this approximation, a rolling-shutter image captured at $\tilde{t}_k$ with time offset $ \Delta t^o_k$ can be viewed as an global-shutter image captured at $\tilde{t}_k'$ with time offset $ {\Delta t^o_k}'$.\
In the following, we slightly abuse notation by replacing $ {\Delta t^o_k}'$ with $\Delta t^o_k$ and $\tilde{t}_k'$ with $\tilde{t}_k$. 
We also replace $\mathbf{p}_{b_k}^w$ and $\mathbf{R}_w^{b_k}$ in Eq.~\eqref{equ:globalvisualproj} by $\tilde{\mathbf{p}}_{b_k}^w$ and $\tilde{\mathbf{R}}_w^{b_k}$ as they are now evaluated at time instant $\tilde{t}_k$.
To calculate the pose at $\tilde{t}_k$, we use IMU propagation from the pose at $t_k$:
\begin{flalign}
	\begin{split}
	\label{equ:pose_calculation}
	\tilde{\mathbf{p}}_{b_k}^{w} &= \mathbf{p}_{b_k}^{w} + \mathbf{v}^w_{b_k} \Delta t_k^o - \frac{1}{2}\mathbf{g}^{w}  (\Delta t_k^o) ^2 + \mathbf{R}_{b_k}^{w}  \hat{\boldsymbol{\alpha}}_{c_k}^{b_k}, \\
	\tilde{\mathbf{q}}_{b_k}^{w} &= \mathbf{q}_{b_k}^{w} \otimes \hat{\mathbf{q}}_{c_k}^{b_k},
	\end{split}\\
	\begin{split}
	\label{equ:continous_integration}	
	\hat{\boldsymbol{\alpha}}_{c_k}^{b_k}&= \iint_{t \in [t_k, \tilde{t}_k ] }  \mathbf{R}_{t}^{b_k} (\mathbf{a}^{b_t} - \mathbf{b}_a^{b_k} )dt^2, \\
	\hat{\mathbf{q}}_{c_k}^{b_k} &= \int_{t \in [t_k, t_{c_{k}} ] }   \frac{1}{2} \begin{bmatrix}
	- \left \lfloor \boldsymbol{\omega}^{b_t} - \mathbf{b}_{\boldsymbol{\omega}}^{b_k} \right \rfloor_{\times} \ \  & \boldsymbol{\omega}^{b_t} - \mathbf{b}_{\boldsymbol{\omega}}^{b_k}  \\
	-(\boldsymbol{\omega}^{b_t} - \mathbf{b}_{\boldsymbol{\omega}}^{b_k})^T  & 0
	\end{bmatrix} \mathbf{q}_{b_t}^{b_k}  dt
	\end{split}
	\end{flalign}
where $\mathbf{a}^{b_t}$/$ \boldsymbol{\omega}^{b_t}$ is the instant linear acceleration/angular velocity. Since only discrete IMU measurements are available on IMUs, $\hat{\boldsymbol{\alpha}}_{c_k}^{b_k}$ and $\hat{\mathbf{q}}_{c_k}^{b_k} $ in \eqref{equ:continous_integration} are approximately computed using numerical integration (i.e. mid-point integration).

The benefit of our constant-pose approximation is that additional variables, i.e. velocities and the rolling-shutter row time, are not needed for estimation, which leads to a large reduction of computational complexity.

\subsection{Modeling Camera-IMU Time Offset}\label{subsec:offset_model}
From the previous subsection, we see that the time offset $\Delta t^o_k$ is the addition of two parts. The first part is the camera-IMU time offset, which varies smoothly because of the system payload variation and sensor jitter. The second part is the compensated time offset caused by the rolling-shutter effect approximation, which varies smoothly according to the change of local trajectories.
We see $\Delta t^o_k$ as a slowly time-varying quantity and model it as a Gaussian random walk: $\dot{\Delta t}_{k}^o = \mathbf{n}_o$,
where $\mathbf{n}_o$ is zero-mean Gaussian noise with covariance $\mathbf{\Sigma}_{o}$. Since time offsets we optimize are at discrete time instants, we integrate this noise over the time interval between two consecutive frames in the sliding window $[t_{k}, \  t_{k+1}]$: ${\Delta t}_{k+1}^o= {\Delta t}_{k}^o  + \mathbf{n}_k^o,  \mathbf{\Sigma}_{k}^o = \Delta t_k \mathbf{\Sigma}_{o}$, where $\mathbf{n}_k^o$ and $\mathbf{\Sigma}_{k}^o$ are discrete noise and covariance, respectively.
Thus we add $\|{\Delta t}_{k+1}^o- {\Delta t}_{k}^o  \|_{\mathbf{\Sigma}_{k}^o}$ into Eq.~\eqref{equ:MAP} for all consecutive frames. By including constraints between consecutive time offsets, we avoid ``offset jumping'' between consecutive frames.

\subsection{Optimization with Unknown Camera-IMU Time Offset}\label{subsec:modelopt}

Our state vector at time instant $t_k$ reads as
$\mathbf{x}_{b_k} = [\mathbf{p}_{b_k}^w, \mathbf{v}_{b_k}^w, \mathbf{q}_{b_k}^w, \mathbf{b}_a^{b_k}, \mathbf{b}_\omega^{b_k}, \Delta t^o_k]$,
where $\Delta t^o_k$ is the dynamically changing camera-IMU time offset modeling both the approximate compensation of rolling-shutter effects and imperfect sensor synchronization.
With $\Delta t^o_k$, the error state $\mathcal{\delta X}$ for linearization becomes $\mathcal{\delta X} = [\delta\mathbf{p}_{b_k}^w \ \delta\mathbf{v}_{b_k}^w \ \delta\boldsymbol{\theta}_{b_k}^w \ \delta\mathbf{b}_a^{b_k} \ \delta\mathbf{b}_{\omega}^{b_k} \ \delta\mathbf{p}_b^c \ \delta\boldsymbol{\theta}_b^c \ \delta\mathbf{f}_i^w \ \delta \Delta t_{k}^o]$,
where we adopt a minimal error representation for rotations ($\delta\boldsymbol{\theta}_{b_k}, \delta\boldsymbol{\theta}_b^c  \in \mathbb{R}^3$): $
\mathbf{q}_{b_k}^w = \hat{\mathbf{q}}_{b_k}^w \otimes \begin{bmatrix}
\delta\boldsymbol{\theta}_{b_k}^w \\
1
\end{bmatrix}, \ \
\mathbf{q}_b^c = \hat{\mathbf{q}}_b^c \otimes \begin{bmatrix}
 \delta\boldsymbol{\theta}_b^c \\
1
\end{bmatrix}.$
Other error-state variables $\delta\mathbf{p}_{b_k}^w$, $\delta\mathbf{v}_{b_k}^w$, $\delta\mathbf{b}_a^{b_k}$, $\delta\mathbf{b}_{\omega}^{b_k}$, $\delta\mathbf{p}_b^c$, $\delta\mathbf{f}_i^w$, and $\delta \Delta t_{k}^o$ are standard additive errors.
After introducing $\delta \Delta t_{k}^o$,
the Jacobian $\mathbf{H}_k$ of the inertial residual function in Eq. \eqref{equ:MAP_error} remains the same as before,
while the Jacobian $\mathbf{H}^i_k$ of the visual residual function needs to be reformulated.
Denoting $\hat{(\cdot)}$ the states obtained from the last iteration of the nonlinear optimization, the Jacobian $\mathbf{H}^i_k$ can be written as
\begin{flalign}
	&\mathbf{H}^i_k = \frac{\partial r_c}{\partial \mathcal{\delta X}} = \frac{\partial r_c}{\partial \mathbf{f}^{c_k}_i } \frac{\partial \mathbf{f}^{c_k}_i}{\partial \mathcal{\delta X} } = \begin{bmatrix}
	\frac{1}{\hat{z}^{c_k}_i} \  & 0 \  & -\frac{\hat{x}^{c_k}_i}{\hat{z}^{c_k}_i} \\
	0 \  & \frac{1}{\hat{z}^{c_k}_i} \  & -\frac{\hat{y}^{c_k}_i}{\hat{z}^{c_k}_i}
	\end{bmatrix} \mathbf{J}, \nonumber\\
	&\mathbf{J} = [-\hat{\mathbf{R}}^c_b\tilde{\mathbf{R}}^{b_k}_w  \  -\hat{\mathbf{R}}^c_b\tilde{\mathbf{R}}^{b_k}_w \Delta \hat{t}_k^o  \ \  \hat{\mathbf{R}}^c_b\tilde{\mathbf{R}}^{b_k}_w  \left \lfloor \hat{\mathbf{f}}^w_i - \tilde{\mathbf{p}}_{b_k}^w  \right \rfloor_{\times} \  \mathbf{0}_{3 \times 6} \ \ \hat{\mathbf{R}}^c_b \ \left \lfloor \hat{\mathbf{f}}^{c_k}_i \right \rfloor_{\times} \  \hat{\mathbf{R}}^c_b\tilde{\mathbf{R}}^{b_k}_w \ \mathbf{J}_{\delta \Delta t}], \nonumber\\
	&\mathbf{J}_{\delta \Delta t} = \hat{\mathbf{R}}^c_b( \left \lfloor \boldsymbol{\omega}^{b_k} \right \rfloor_{\times} \tilde{\mathbf{R}}^{b_k}_w ( \hat{\mathbf{f}}^w_i - \tilde{\mathbf{p}}_{b_k}^w  ) + \tilde{\mathbf{R}}^{b_k}_w \mathbf{v}^w_{b_k} + \mathbf{g}^w \Delta \hat{t}_k^o ),\nonumber
\end{flalign}
where $\left \lfloor \cdot \right \rfloor_{\times}$ denotes the skew symmetric matrix of a vector.
Recall that the position $\tilde{\mathbf{p}}_{b_k}^w$ and rotation $\tilde{\mathbf{R}}_w^{b_k}$ in the Jacobian is computed at time instant $\tilde{t}_k$ instead of $t_k$,
which depends on variable $\Delta \hat{t}_k^o$ and varies in each iteration of the optimization.
Besides, the computation of the feature position $\hat{\mathbf{f}}^{c_k}_i$ projected on the $k$-th frame depends on $\tilde{\mathbf{p}}_{b_k}^w$ and $\tilde{\mathbf{R}}_w^{b_k}$ as shown in Eq. \eqref{equ:globalvisualproj}, which also needs to be recomputed when $\Delta \hat{t}_k^o$ changes.
We in the next section present a novel algorithm based on dynamic programming and red-black tree to efficiently compute $\tilde{\mathbf{p}}_{b_k}^w$ and $\tilde{\mathbf{R}}_w^{b_k}$ as the estimated time offset $\Delta \hat{t}_k^o$ varies during the iterations.
\begin{figure}[t]
	\centering
	\includegraphics[width=0.8\textwidth]{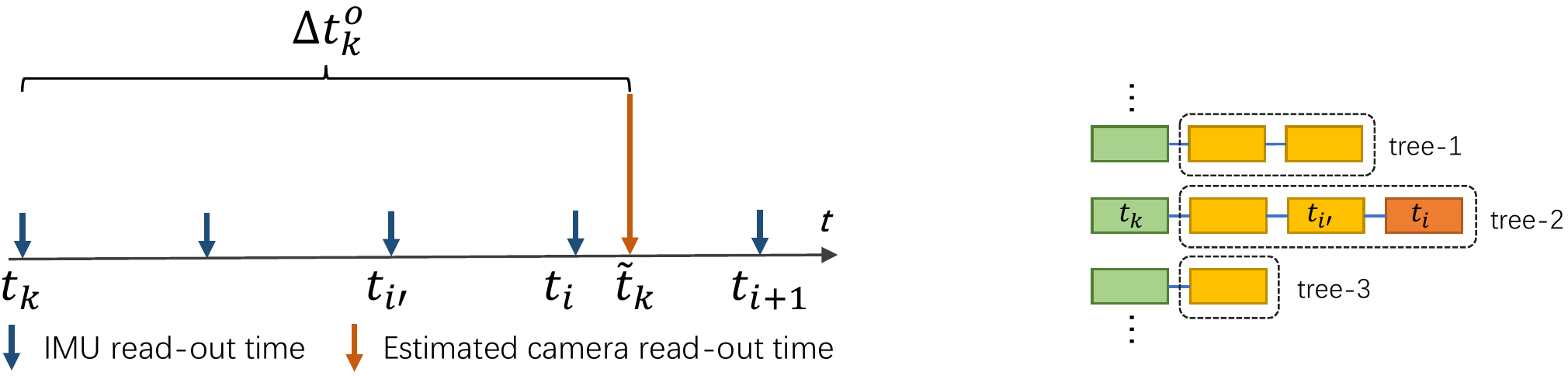}
	\caption{Computing the pose at $\tilde{t}_k$ from the pose at $t_k$ is decomposed into two steps (left): firstly, compute the pose at $t_{i}$, where $t_{i}$ is the closest IMU measurement time instant to $\tilde{t}_k$ and smaller than $\tilde{t}_k$; secondly, compute the pose at $\tilde{t}_k$ based on the pose at $t_{i}$. We design an algorithm based on dynamic programming and red-black tree (right) for efficient indexing to accelerate the first step.}
	\label{fig:preintegration}
\end{figure}
\subsection{Efficient IMU Integration over Variable-Length Time Intervals}\label{subsec:efficientalgo}
With a naive implementation, the computation of $\hat{\boldsymbol{\alpha}}_{c_k}^{b_k}$ and $\hat{\mathbf{q}}_{c_k}^{b_k} $ in Eq. \eqref{equ:continous_integration} between $t_k$ and $\tilde{t}_k$ needs to be recomputed each time the offset $\Delta t_k^o$ changes during the optimization.
In order to reuse the intermediate integration results, we decompose the integration into two steps (Fig.~\ref{fig:preintegration}):
firstly, compute the integration between $t_k$ and $t_i$; secondly, compute the integration between $t_i$ and $\tilde{t}_k$.
Here, without loss of generality, we assume $t_i$ is the closest IMU read-out time instant before $\tilde{t}_k$.
The decomposition makes the results in the first step reusable since the integration is computed over variable yet regular time intervals.
We design an algorithm based on discrete dynamic programming \cite{knuth1998} and red-black tree \cite{knuth1998} to perform efficient integration over the variable-length time intervals.

Specifically, we build a table (implemented as a red-black tree \cite{knuth1998}) for each time instant $t_k$ to store the intermediate results of IMU integration starting from $t_k$ (right part of Fig.~\ref{fig:preintegration}).
Each node in the tree stores a key-value pair, where the key is an IMU read-out time $t_i$ and the value is the integration from $t_k$ to $t_i$ computed using Eq.~\eqref{equ:continous_integration}.
Note that this integration is independent of the pose and velocity at $t_k$, so the stored results do not need to be updated when the pose and velocity in Eq.~\eqref{equ:pose_calculation} as $t_k$ change.
During the optimization, each time when we need to compute the integration from $t_k$ to $t_i$ (recall that $t_i$ is variable according to $\tilde{t}_k$),
we first try to search in the tree at $t_k$ to query the integration results for $t_i$.
If the query failed, we then instead search if there exists a record for another IMU read-out time $t_{i'}$ such that $t_{i'} < t_i$.
If multiple records exist, we select the maximum $t_{i'}$ (which is closest to $t_i$),
and compute the integration from $t_k$ to $t_i$ based on the retrieved record at $t_{i'}$ and IMU measurements from $t_{i'}$ to $t_i$.
During the process, each time a new integration is computed, the result is stored into the tree for future retrieval.

\section{Uncertainty-Aware Initialization}\label{sec:initalization}
We initialize our VIO system by first running a vision-only bundle adjustment on $K$ consecutive frames.
Suppose that the $K$ consecutive rotations and positions obtained from vision-only bundle adjustment are $[\mathbf{p}_{c_0}^{c_0} \ \mathbf{R}_{c_0}^{c_0} \ ... \  \mathbf{p}_{c_{K-1}}^{c_0} \ \mathbf{R}_{c_{K-1}}^{c_0}]$.
The goal of the initialization is to solve initial velocities $[\mathbf{v}_{b_k}^{b_0} \ ... \ \mathbf{v}_{b_{K-1}}^{b_0}]$,
gravity vector $\mathbf{g}^{c_0}$, and metric scale $s$.
The initializations in \cite{tong17,orbimu} solve for the unknowns by minimizing the least squares of the $L_2$ errors between the poses obtained from vision-only bundle adjustment and the poses obtained from IMU integration.
These methods do not take into account the uncertainties introduced by IMU noise during the IMU integration,
which can cause failures of the initialization when the camera-IMU time offset is unknown (see Sect. \ref{seq:expmobiledata}).

We employ a different method to perform uncertainty-aware initialization by incorporating the covariances obtained during the IMU integration.
The IMU propagation model for the $K$ initialization images can be written as
\begin{flalign}
\label{equ:init}
\begin{split}
 \mathbf{p}_{b_{k+1}}^{c_0}  &= \mathbf{p}_{b_{k}}^{c_0} + \mathbf{R}_{b_k}^{c_0} \mathbf{v}_{b_k}^{b_k}\Delta t_k  - \frac{1}{2} \mathbf{g}^{c_0} \Delta t_k^2 + \mathbf{R}_{b_k}^{c_0} \hat{\boldsymbol{\alpha}}_{k+1}^{k}, \\
 \mathbf{R}_{b_{k+1}}^{c_0} \mathbf{v}_{b_{k+1}}^{b_{k+1}}  &= \mathbf{R}_{b_k}^{c_0} \mathbf{v}_{b_k}^{b_k}  -  \mathbf{g}^{c_0} \Delta t_k + \mathbf{R}_{b_k}^{c_0} \hat{\boldsymbol{\beta}}_{k+1}^{k},
 \end{split}
\end{flalign}
where $\mathbf{p}_{b_{k}}^{c_0} = s \mathbf{p}_{c_{k}}^{c_0} - \mathbf{R}_{b_k}^{c_0} \mathbf{p}^b_{c} $ and $\mathbf{R}_{b_k}^{c_0} = \mathbf{R}_{c_k}^{c_0} \mathbf{R}_{b}^{c} $.
We set the camera-IMU extrinsics $\mathbf{p}^b_{c} $ and $\mathbf{R}^b_{c} $ to a zero vector and an identity matrix, respectively.
Instead of minimizing the $L_2$ norm of the measurement errors, we minimize the Mahalanobis norm of the measurement errors as:
{\scriptsize \begin{flalign}
\label{equ:intialization}
\underset{\mathcal{X}} {\min}& \sum_{\hat{\mathbf{z}}_{k+1}^{k} \in S_i} \Bigg\| \begin{bmatrix}
\mathbf{R}^{b_k}_{c_0} ( s \mathbf{p}_{c_{k+1}}^{c_0} - \mathbf{R}_{b_{k+1}}^{c_0} \mathbf{p}^b_{c} -  s \mathbf{p}_{c_{k}}^{c_0} + \mathbf{R}_{b_k}^{c_0} \mathbf{p}^b_{c} + \frac{1}{2} \mathbf{g}^{c_0} \Delta t_k^2 ) - \mathbf{v}_{b_k}^{b_k}\Delta t_k - \hat{\boldsymbol{\alpha}}_{{k+1}}^{k} \\
\mathbf{R}^{b_k}_{c_0} ( \mathbf{R}_{b_{k+1}}^{c_0} \mathbf{v}_{b_{k+1}}^{b_{k+1}} +  \mathbf{g}^{c_0} \Delta t_k ) - \mathbf{v}_{b_k}^{b_k} -  \hat{\boldsymbol{\beta}}_{k+1}^{k} \\
\end{bmatrix} \Bigg\|_{\mathbf{\Sigma}_{k+1}^{k} }^2, \nonumber
\end{flalign}}\\
where $\hat{\mathbf{z}}_{k+1}^{k} = \{\hat{\boldsymbol{\alpha}}_{k+1}^{k}, \hat{\boldsymbol{\beta}}_{k+1}^{k}\}$ and $\mathbf{\Sigma}_{k+1}^{k}$ is the covariance matrix computed during IMU integration \cite{forster15}.
Note that the covariance $\mathbf{\Sigma}_{k+1}^{k}$ in our formulation models the uncertainty of the IMU measurements.
The resulting problem is a weighted least squares problem and can be solved efficiently.
After solving the problem, we project the solved variables into the world coordinate.
Other variables like accelerator biases, gyrocope biases, and time offsets are initialized as zeros.

\section{Experiments}\label{sec:experiments}
We compare our approach to state-of-the-art monocular VIO systems: OKVIS \cite{leuRSS2013} and VINS-Mono \cite{qin2017vins}.
Loop closure in VINS-Mono is disabled for a fair comparison.
The number of keyframes and non-keyframes in the optimization window of OKVIS and our approach are set to be 8 and 3 respectively.
The sliding window size in VINS-Mono is set to be 11.

\subsection{Performance on The Euroc Dataset}\label{sec:expeurodata}
\begin{figure}[t]
	\centering
	\subfloat[Rot. errors without offsets]{{\includegraphics[width=0.41\textwidth]{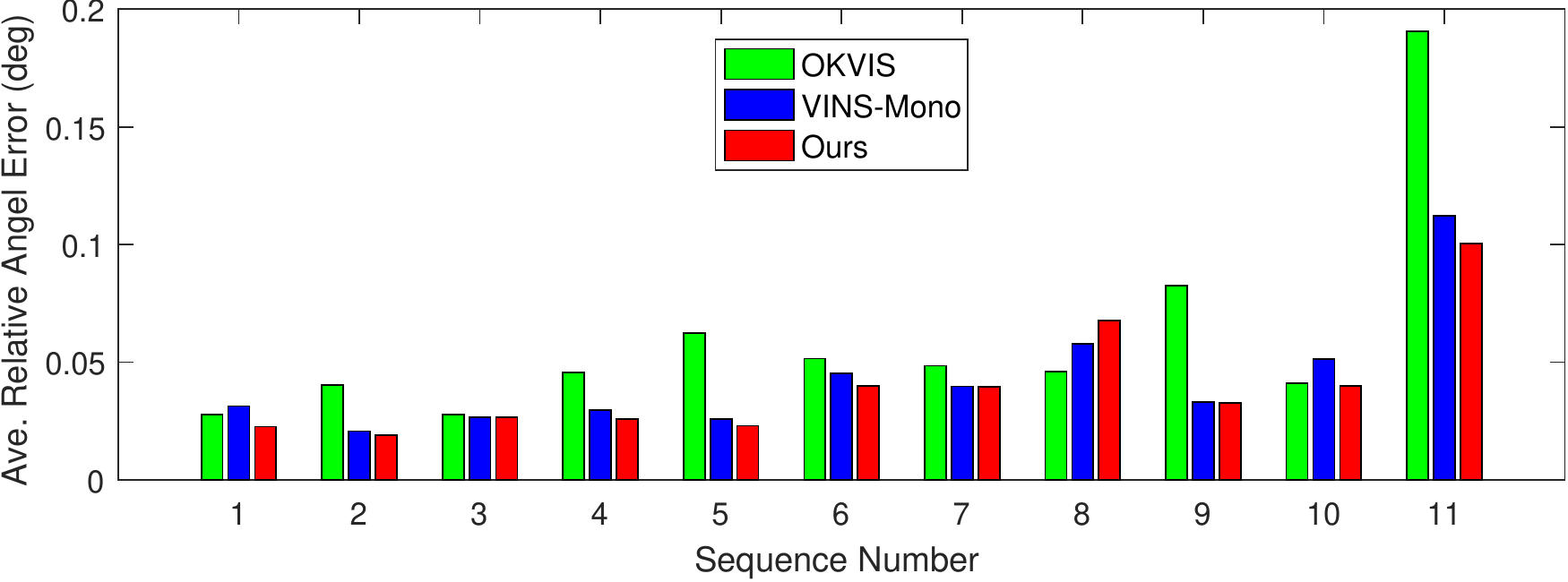} }} \ \ \ \ \
	\subfloat[Trans. errors without offsets]{{\includegraphics[width=0.41\textwidth]{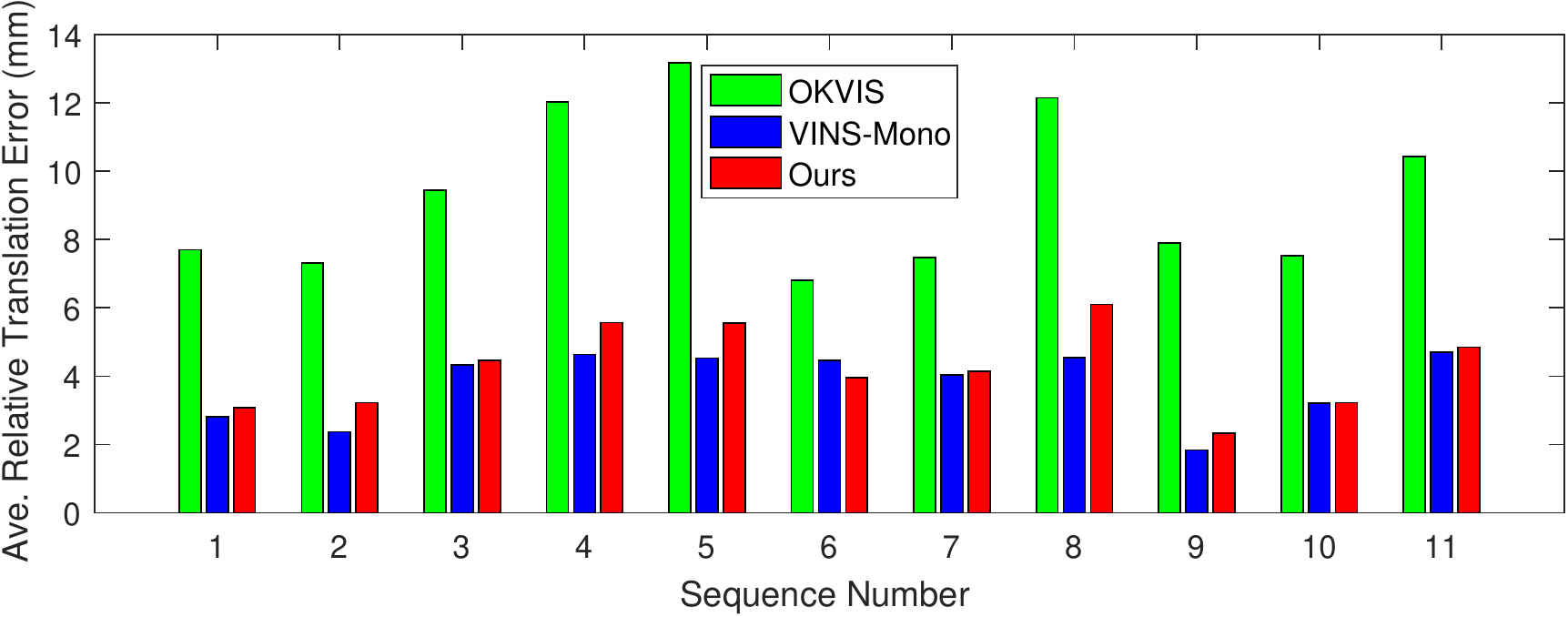} }}  \\
	\subfloat[{Rot. errors with 30 ms offset} ]{{\includegraphics[width=0.41\textwidth]{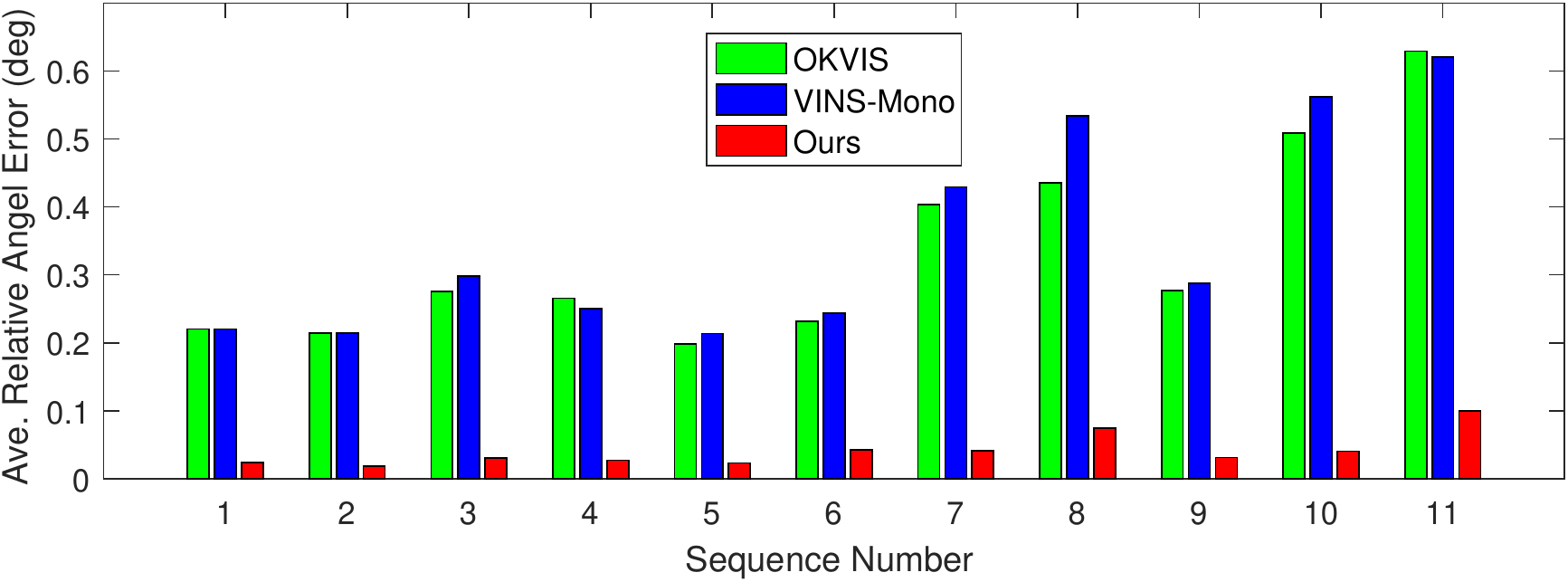} }} \ \ \ \ \
	\subfloat[{Trans. errors with 30 ms offset}]{{\includegraphics[width=0.41\textwidth]{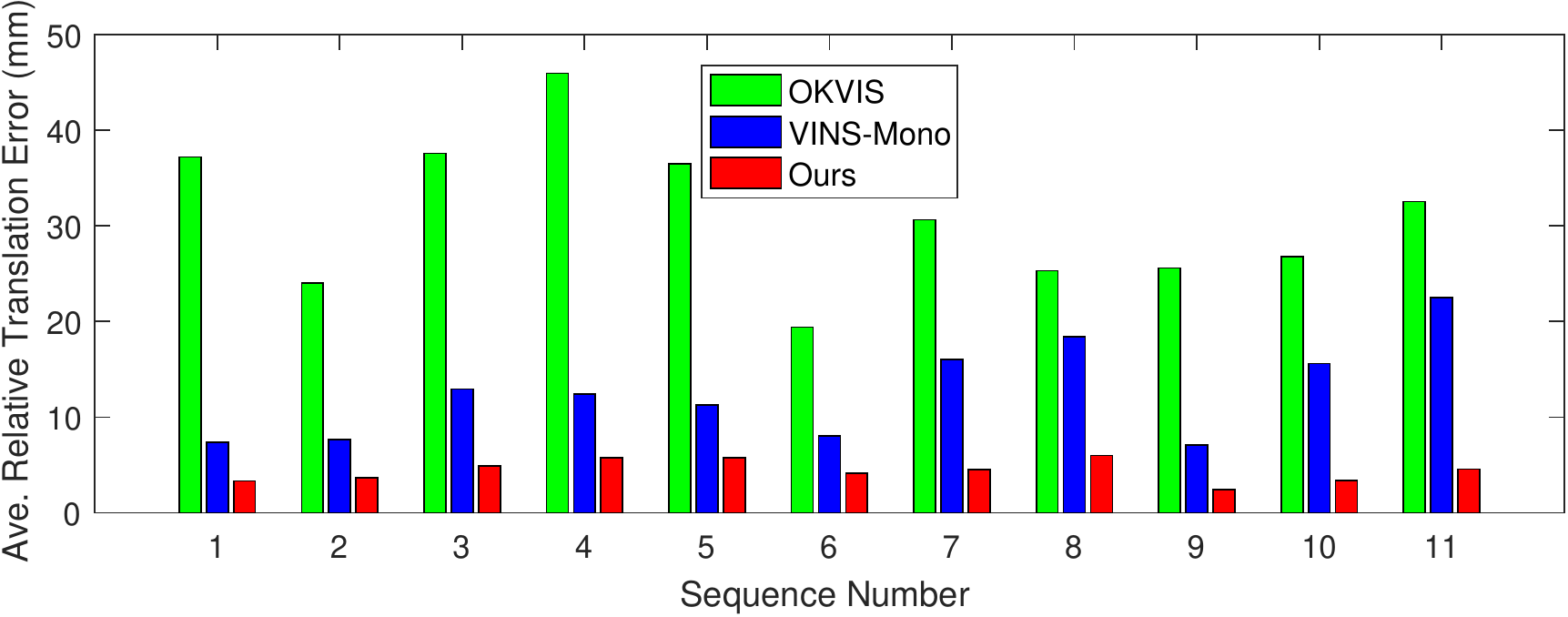} }}  \\
	\subfloat[{Rot. errors with 60 ms offset}]{{\includegraphics[width=0.41\textwidth]{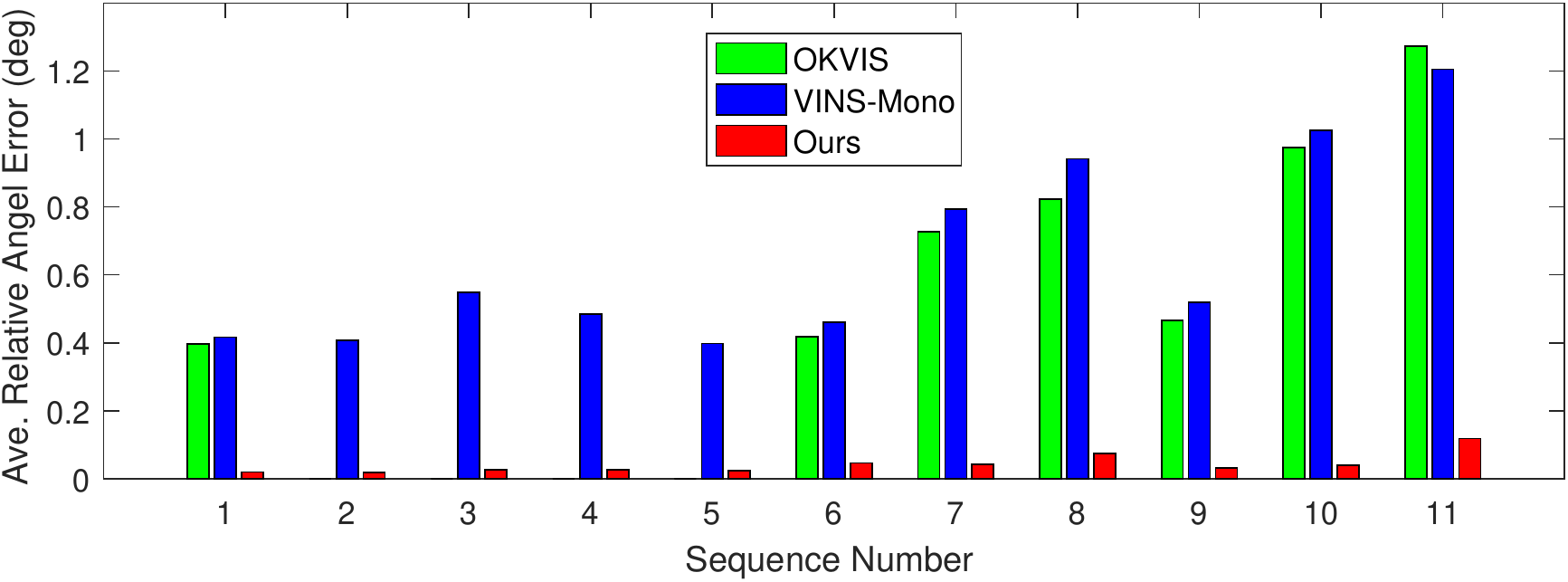} }} \ \ \ \ \
	\subfloat[{Trans. errors with 60 ms offset}]{{\includegraphics[width=0.41\textwidth]{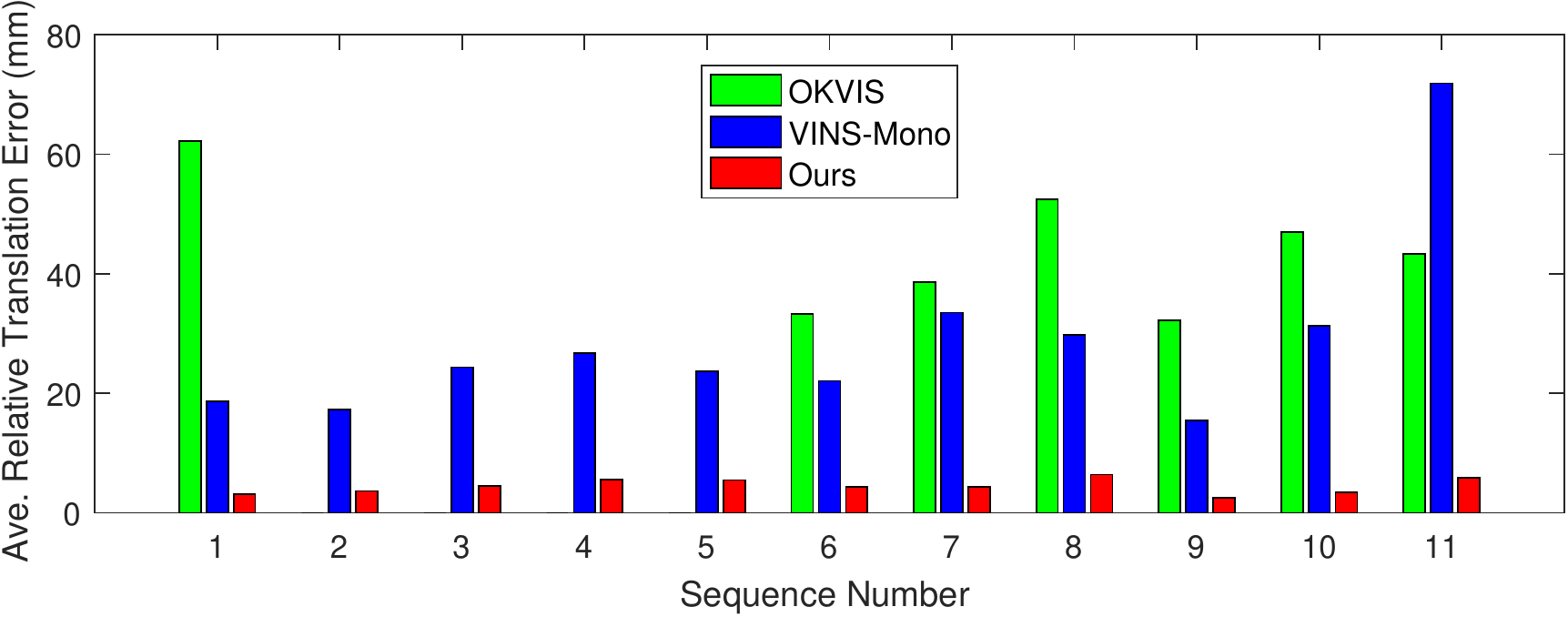} }}
	\caption{The relative rot. and trans. errors with different time offsets on Euroc.}
	\label{fig:offset}
\end{figure}
\begin{figure}[t]
	\centering
	\subfloat[{Avg. relative rot. error (deg)}]{{\includegraphics[width=0.41\textwidth]{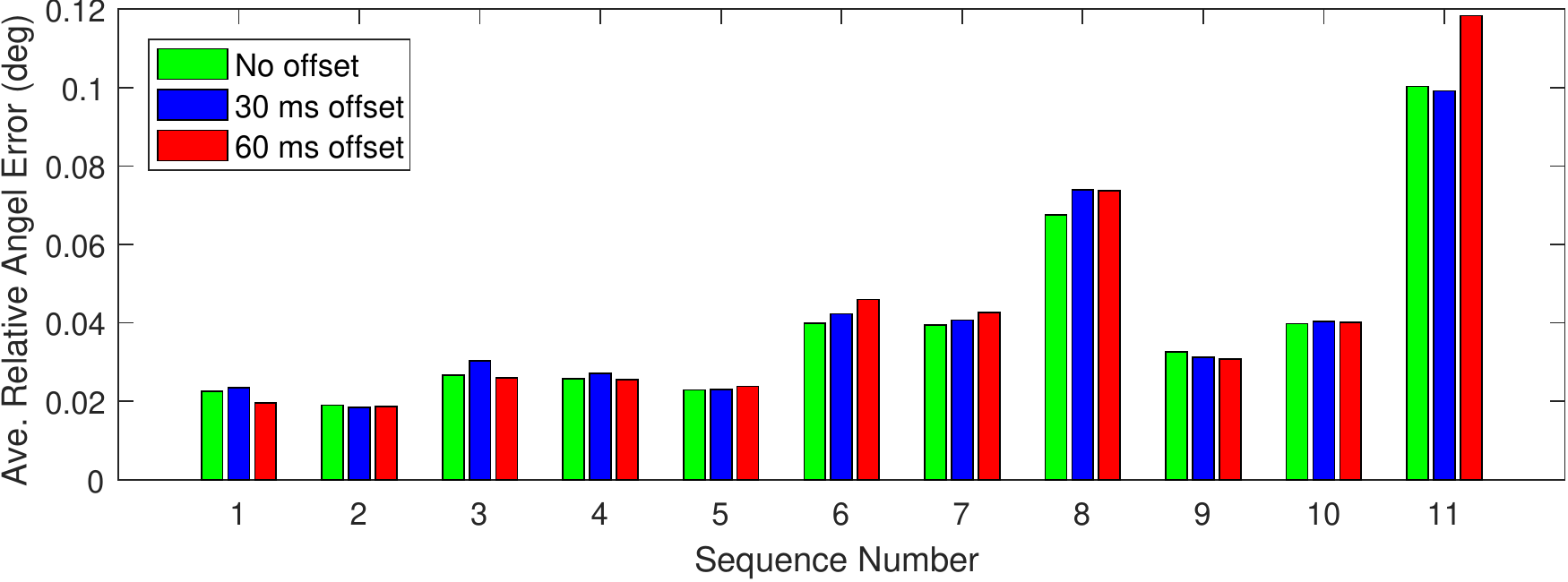} }} \ \ \ \ \
	\subfloat[{Avg. relative trans. error (m)}]{{\includegraphics[width=0.41\textwidth]{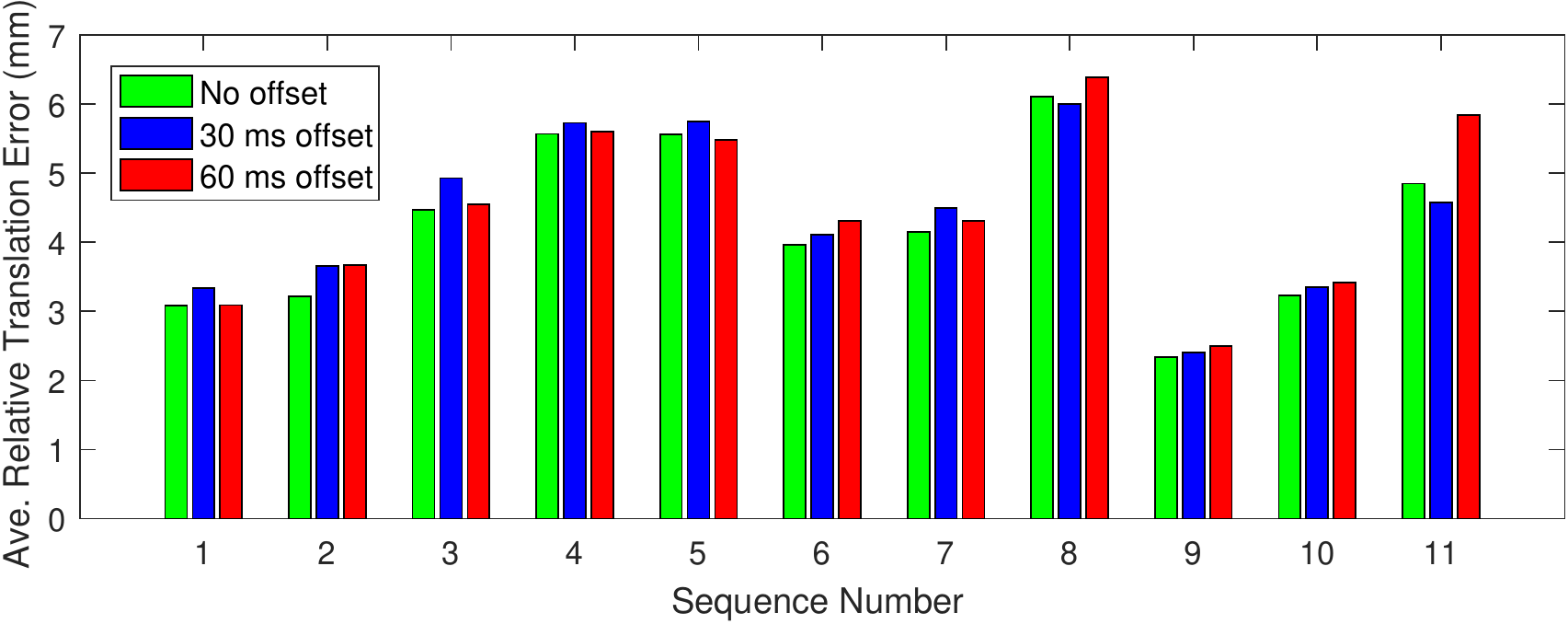} }}  \\
	\caption{The average relative rot. and trans. errors of our approach with different time offsets on Euroc.}
	\label{fig:offset_ours}
\end{figure}

Sequences of the Euroc dataset consist of synchronized global-shutter stereo images (only mono/left images are used) and IMU measurements. The complexity of these sequences varies regarding trajectory length, flight dynamics, and illumination conditions. Ground truth poses are obtained by Vicon motion capture system.
For presentation, we use numbers to denote sequence names: 1 $\sim$ 5 for MH\_01\_easy $\sim$ MH\_05\_difficult, 6 $\sim$ 8 for V1\_01\_easy $\sim$ V1\_03\_difficult, 9 $\sim$ 11 for V2\_01\_easy $\sim$ V2\_03\_difficult.

\noindent \textbf{Significance of Online Temporal Camera-IMU Calibration}
Two error metrics are used for tracking accuracy evaluation: the average relative rotation error (deg) and the average relative translation error (m) \cite{Geiger2012CVPR}. Approaches are compared under different simulated time offsets between visual and inertial measurements: no time offsets; 30~ms offset; 60~ms offset. Fig.~\ref{fig:offset} shows the comparison results. When visual measurements and inertial measurements are synchronized, the average relative rotation/translation error between VINS-Mono and our approach are similar, while OKVIS performs the worst. As the time offset increases, VINS-Mono and OKVIS perform worse and worse due to the lack of camera-IMU time offset modeling.
Our estimator achieves the smallest tracking error when there exists a time offset.
OKVIS fails to track in sequence 2 $\sim$ 5  when the time offset is set to be 60~ms. We have tested different approaches under larger time offsets, such as 90~ms. However, neither OKVIS or VINS-Mono provides reasonable estimates. We thus omit comparisons for larger time offsets. We also illustrate the tracking performance of our approach under different time offsets in Fig.~\ref{fig:offset_ours}. We see that, by modeling the time offset properly, there is no significant performance decrease as time offset increases. Fig.~\ref{fig:euroc} gives a visual comparison on trajectories estimated by our approach and VINS-Mono on the Euroc sequence V1\_03\_difficult with a 60~ms time offset. Noticeable `jaggies' are found on the trajectory estimated by VINS-Mono.

\begin{figure}[t]
	\centering
\begin{minipage}[]{0.25\textwidth}
	\includegraphics[width=1.1\textwidth]{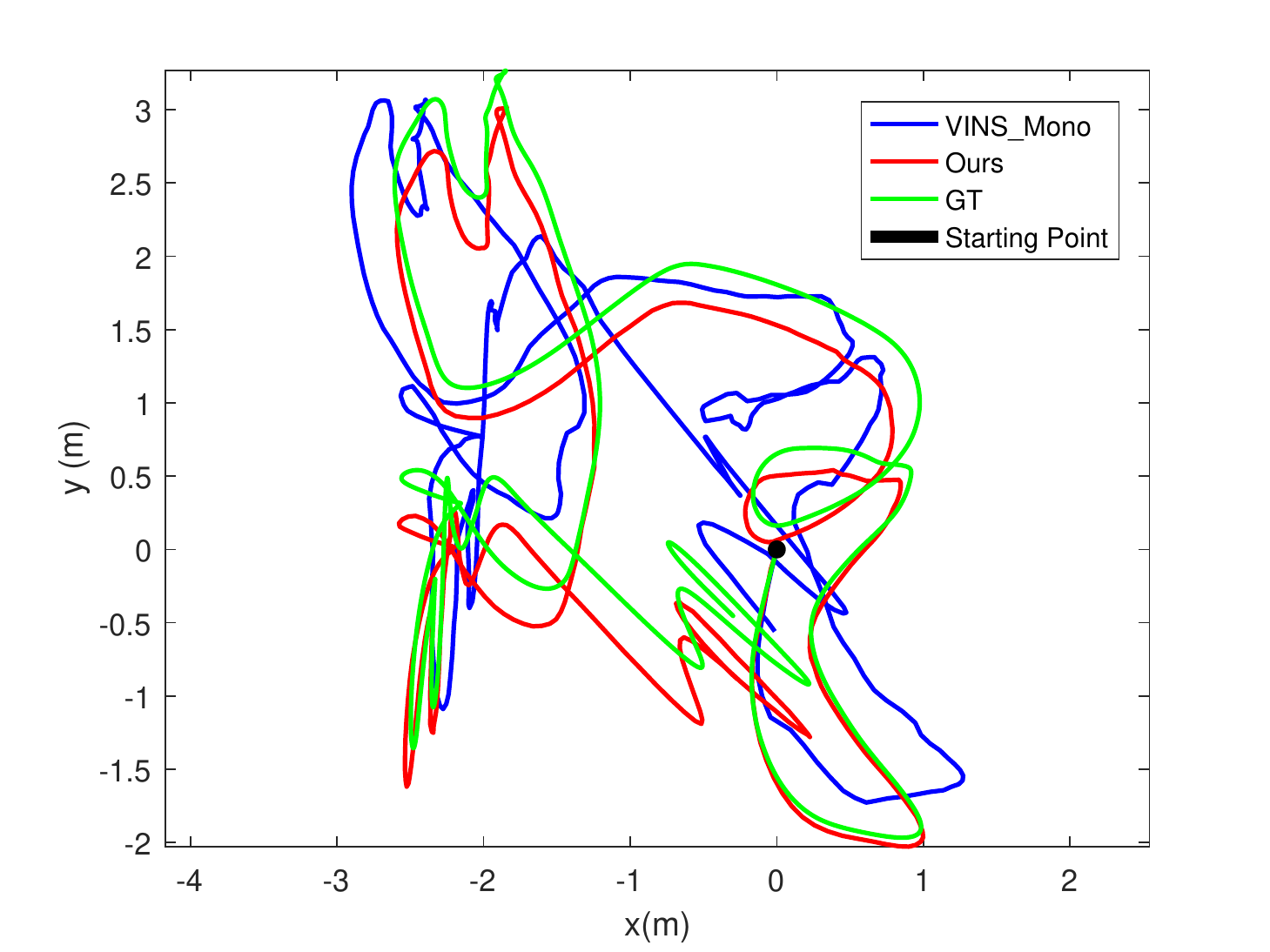}
	\caption{Comparison on the Euroc V1\_03\_difficult with a 60~ms time offset.}
	\label{fig:euroc}
\end{minipage} \ \
\begin{minipage}[]{0.7\textwidth}
	\includegraphics[width=0.5\textwidth]{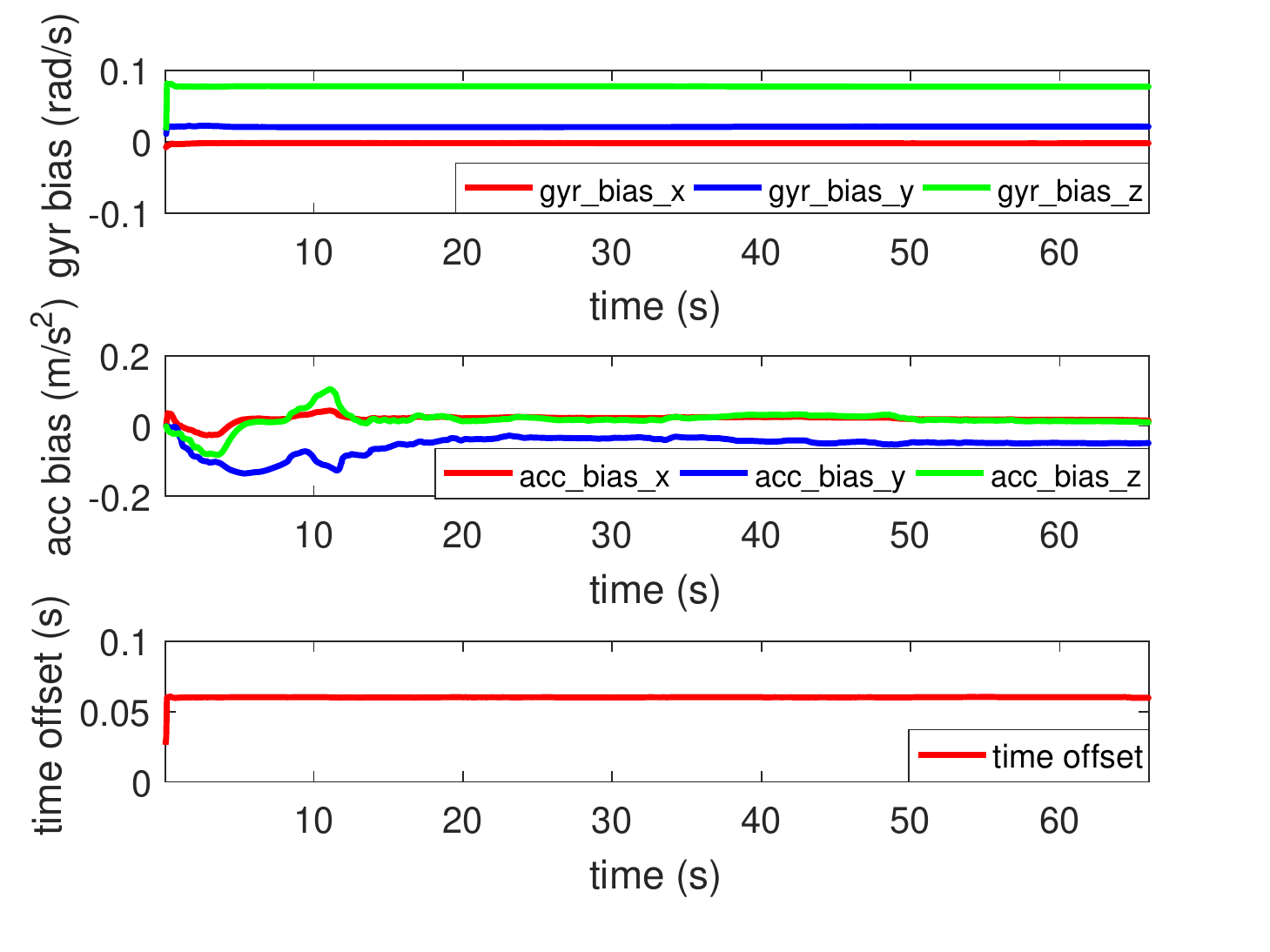}
	\includegraphics[width=0.5\textwidth]{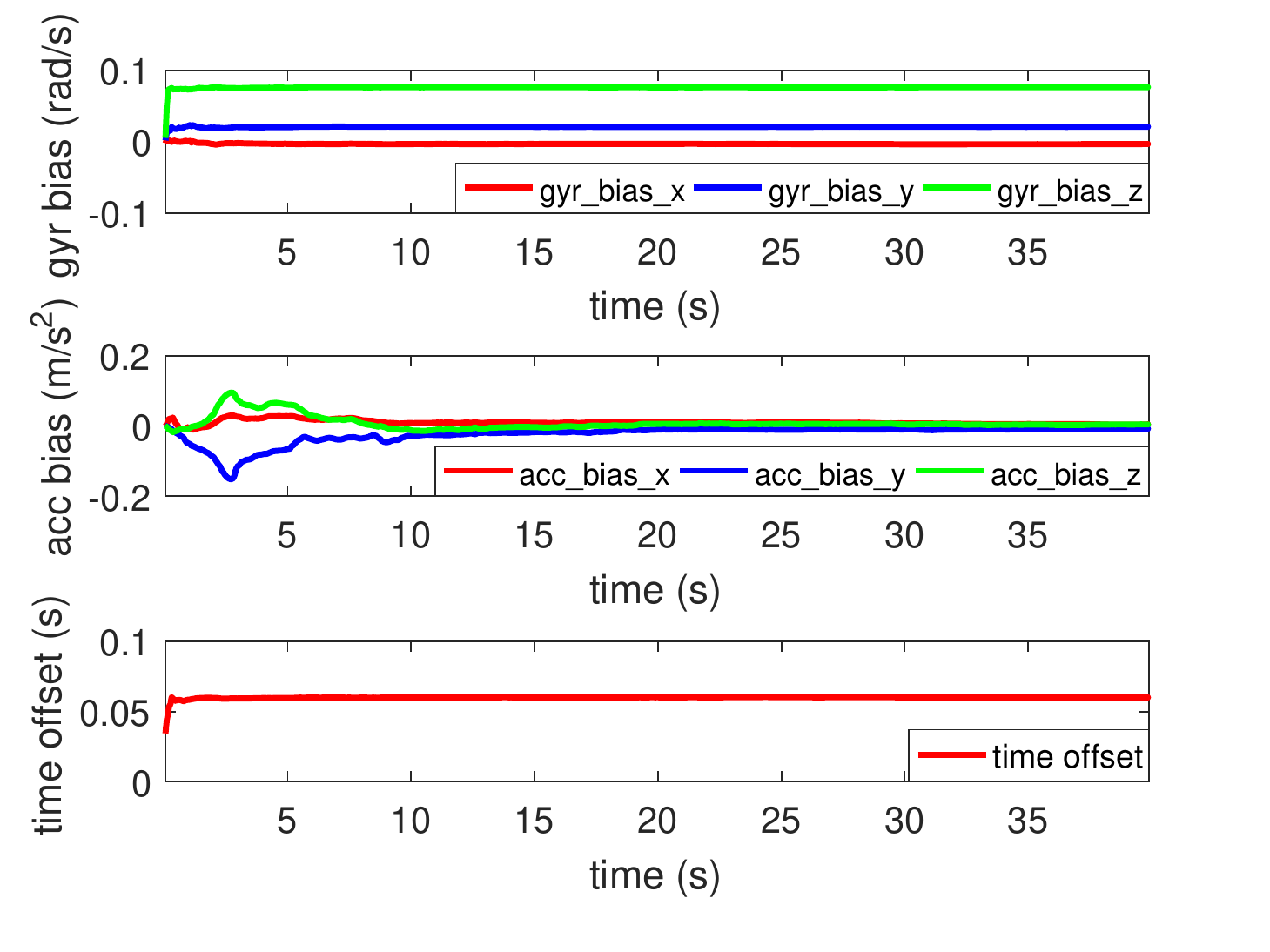}
	\caption{The convergence study of variables in our estimator on the Euroc MH\_03\_medium (left) and V1\_02\_medium (right) with a 60~ms time offset.}
	\label{fig:convergence}
\end{minipage}
\end{figure}

\noindent \textbf{Parameter Convergence Study}
Another aspect we pay attention to is whether estimator parameters converge or not.
We analyze our results on the sequence MH\_03\_medium and sequence V1\_02\_medium with a simulated 60~ms time offset. Estimated biases and time offsets w.r.t. time are plotted in Fig.~\ref{fig:convergence}. Both gyroscope biases and time offsets converge quickly. On the one hand, relative rotations are well constrained by visual measurements. They are not related to metric scale estimation. On the other hand, relative rotations are the first-order integration of angular velocities and gyroscope biases, thus they are easy to estimate. Another interesting thing we found is that, the convergence of time offsets is the same as the convergence of gyroscope biases. This means that, time offsets are calibrated mainly by aligning relative rotations from visual constraints and from gyroscope integrations, which is consistent with ideas of offline time offset calibration algorithms \cite{jacovitti93,kelly2010}. Compared to gyroscope biases, acceleration biases converge much more slowly. They are hard to estimate as positions are the second-order integration of accelerations and acceleration biases. Additionally, acceleration measurements obtained from IMUs are coupled with gravity measurements. Abundant rotations collected across time are required to distinguish actual accelerations and gravity vector from measurements.

\subsection{Performance on Mobile Phone Data}\label{seq:expmobiledata}
\begin{figure}[t]
	\subfloat[Slow translation and rotation.]{{\includegraphics[width=0.45\textwidth]{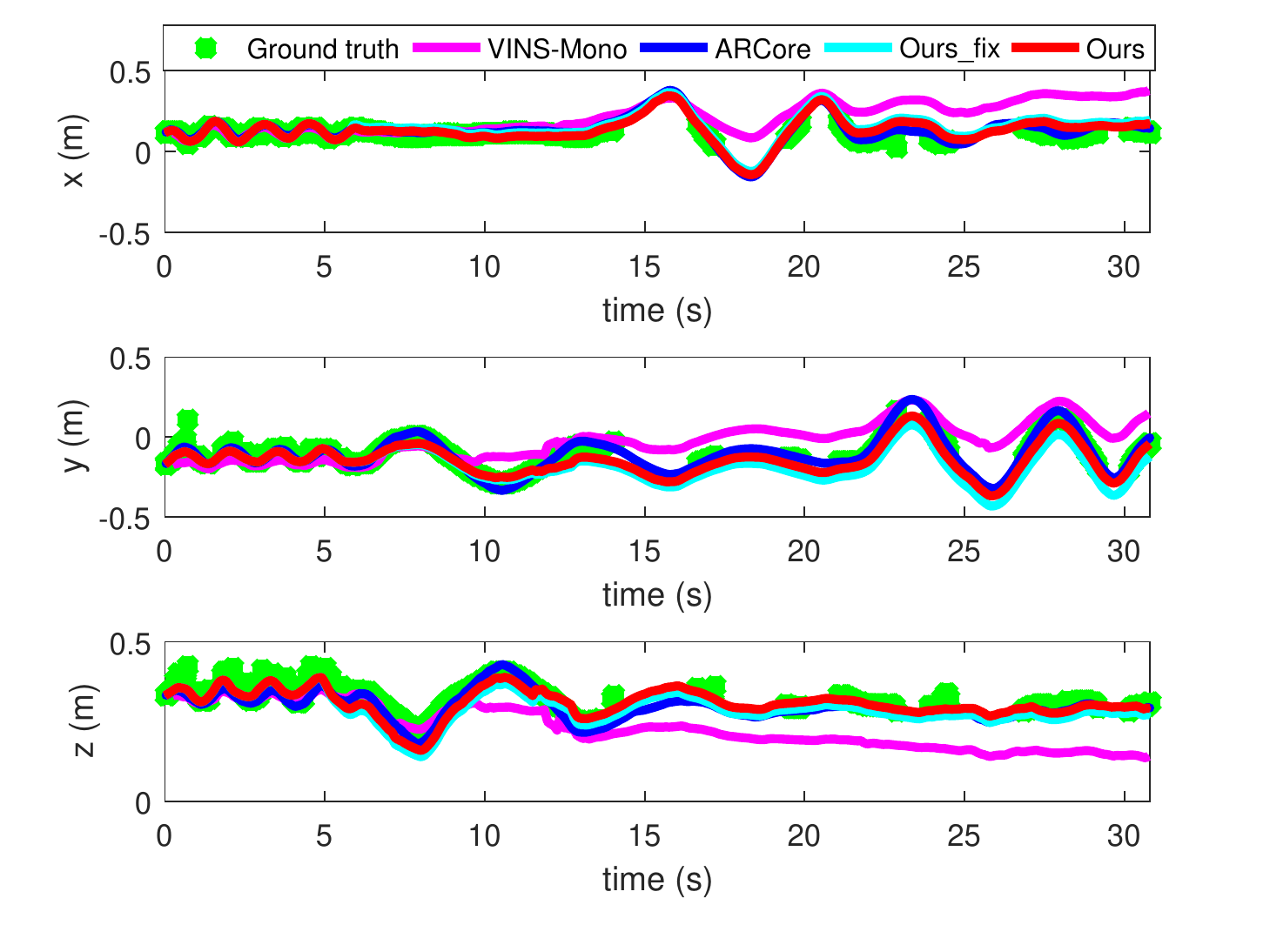} }} \ \ \
	\subfloat[Medium translation and rotation.]{{\includegraphics[width=0.45\textwidth]{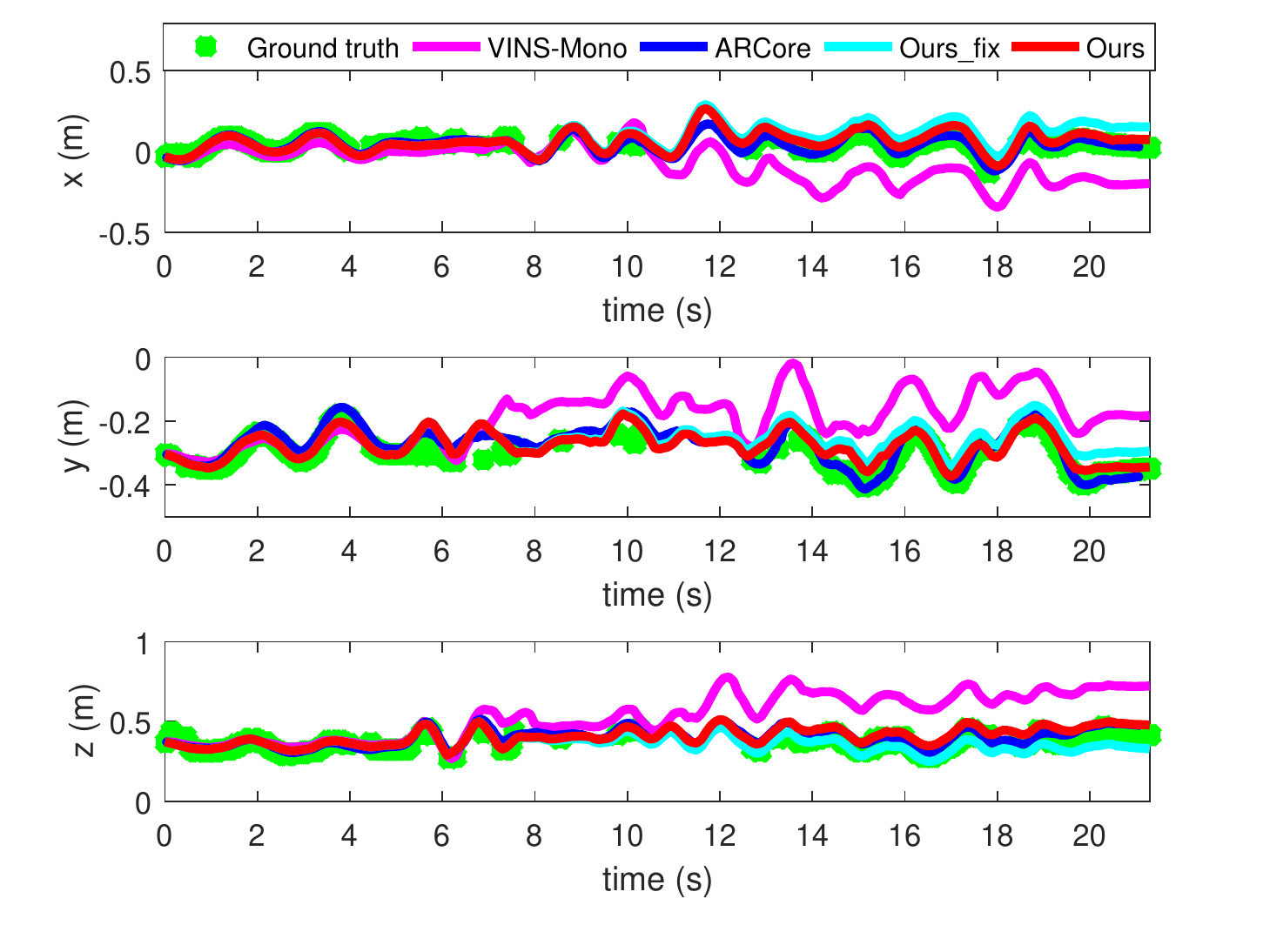} }} \\	
	\subfloat[Fast translation and rotation.]{{\includegraphics[width=0.45\textwidth]{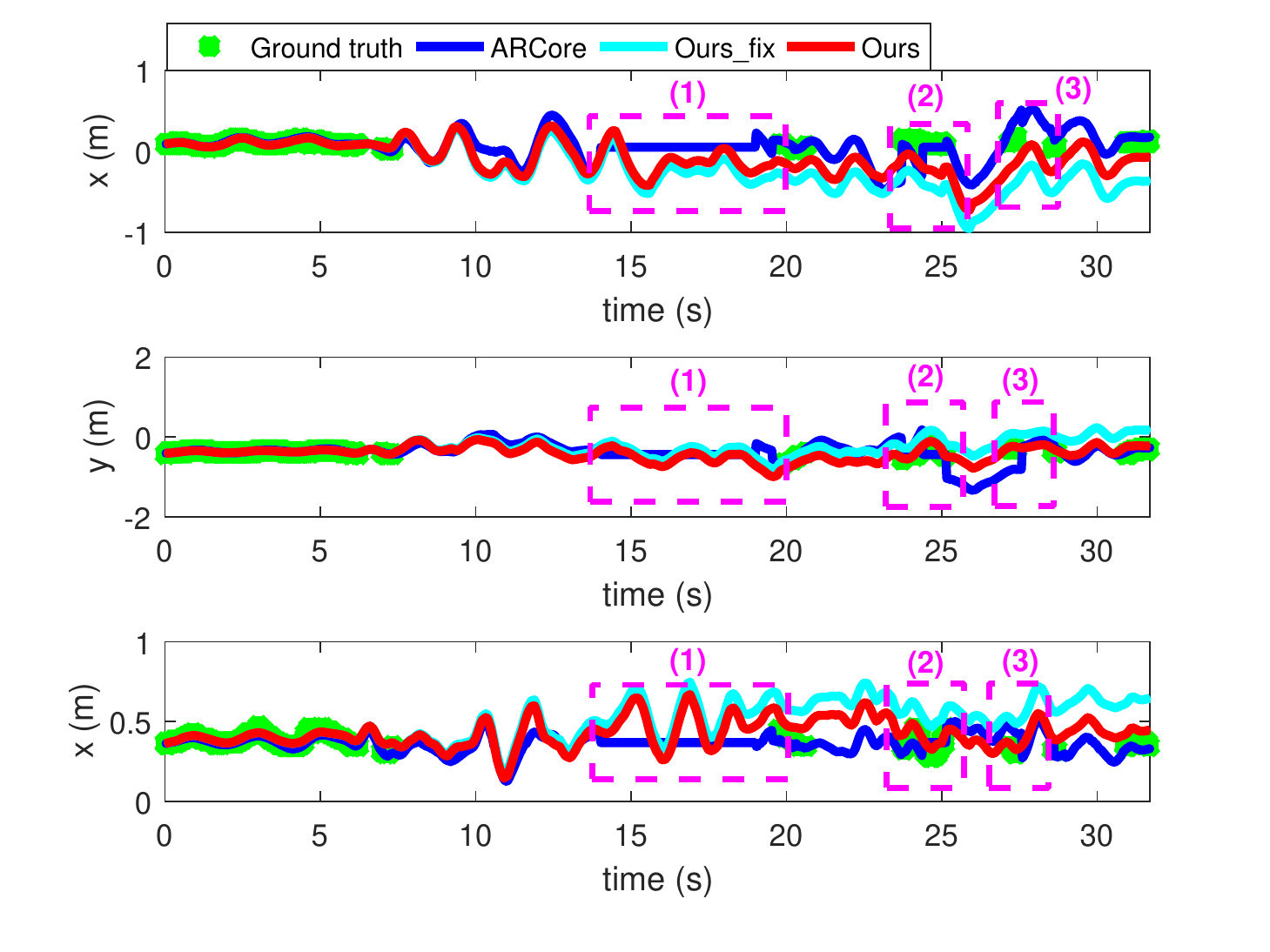} }} \ \ \ \ \ \ \ \ \ \ \ \ \ \ \ \ \ \ \
	\subfloat[Image]{{\includegraphics[width=0.2\textwidth]{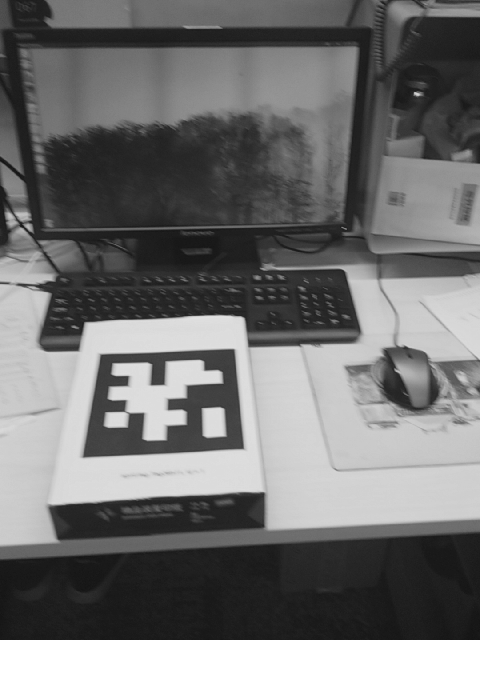}}}
	\caption{The qualitative comparison of tracking accuracy. VINS-Mono fails to track on the fast translation and rotation sequence.}
	\label{fig:comparison}
\end{figure}
One of the applications of VIO is the motion tracking on mobile phones. We collect visual and inertial measurements using Samsung Galaxy S8 with a rolling-shutter camera and unsynchronized sensor measurements.
Google developed ARCore \cite{arcore} is also used for comparison.
After recording the data, we run our system on the Samsung Galaxy S8. It takes $70\pm10$ms for each nonlinear optimization.\\
\noindent \textbf{Qualitative Comparison of Tracking Accuracy}
We use AprilTag \cite{wang2016iros} to obtain time-synchronized and drift-free ground truths (Fig.~\ref{fig:comparison} (d)). We detect tags on captured images. If tags are detected, we calculate ground truth camera poses via P3P \cite{p3p} using tag corners. Since VINS-Mono performs better than OKVIS, we only include tracking results from VINS-Mono for the following comparison.
We test our approach with two conditions: 1)`Ours-fix': the variable time-delay estimation is switched off, i.e. is replaced with a single time offset, to account for the camera-IMU synchronization; 2)`Ours': the variable time-delay is enabled to account for both the camera-IMU synchronization and the rolling-shutter effect approximation.
Three sequences are recorded around an office desk.
These datasets are increasingly difficult to process in terms of motion dynamics: slow, medium, and fast translation and rotation (Fig.~\ref{fig:comparison}).
To align coordinates of different approaches for comparison, we regularly move the mobile phone at the beginning of recorded sequences (see 0\--5s in (a), 0\--4s in (b), and 0\--6s in (c)).
VINS-Mono performs worst among all datasets, as it does not model time offsets between visual and inertial measurements.
Our approach with condition `Ours-fix' performs worse than the one with condition `Ours'.
The tracking accuracy of our approach is comparable to that of ARCore in small and medium motion settings. While for the fast translation and rotation sequence, where part of captured images are blurred, our approach exhibits better tracking robustness compared to ARCore. ARCore loses tracks in time periods within dashed boxes (1)(2)(3). This is because ARCore is based on EKF \cite{HesKotBow1402}, which requires good initial guesses about predicted states. If pose displacements are large and feature correspondences are not abundant because of image blur, ARCore may fail. Conversely, our nonlinear optimization approach iteratively minimizes errors from sensor measurements, which treats the underlying non-linearity better and are less sensitive to initial guesses. Note that, there is a loop closing module in ARCore for pose recovery and drift correction. The final position drift of ARCore is thus smallest. However, loop closing is not the focus of this work. \\
\noindent \textbf{Qualitative Comparison of Estimator Initialization}
To evaluate the significance of our initialization method, we compare it with the state-of-the-art visual-inertial initialization method \cite{orbimu}. We record 20 testing sequences on Samsung Galaxy S8 with medium translation and rotation. Each testing sequence lasts for about 30 seconds. We use the first 2-second sensor measurements to do the initialization. After the initialization is done, we use the visual-inertial estimator proposed in this work estimate camera poses. We then calculate an optimal scale factor by aligning the estimated trajectory with the trajectory reported by ARCore by a similarity transformation \cite{horm87}. Although there is inevitably pose drift in ARCore estimates, this drift is not significant compared to the scale error caused by improper initializations. If the estimator fails to track in any time intervals of the testing sequences or the calculated scale error is larger than 5\%, we declare the initialization as failures. For a fair comparison, we use the same relative poses obtained by visual sfm \cite{murAcceptedTRO2015} as the initialization input of \cite{orbimu} and that of our approach. We find that 14 successful trials out of 20 (70\%) using initialization proposed in \cite{orbimu}, while 18 successful trials out of 20 (90\%) using our initialization method. The successful initialization rate using our approach is higher than using initialization in \cite{orbimu}. We study the two testing sequences where our initialization fails. We find that time offsets between visual and inertial measurements in these two sequences are larger than 100~ms (this can be obtained by enumerating offsets and testing whether the following visual-inertial estimator outputs are close to that of ARCore or not after initialization), causing a big inconsistency when aligning the visual and inertial measurements. Since the main focus of this work is on the estimator, we are going to handle this failure case thoroughly in the future work.

\section{Conclusions}
\label{sec:conclusions}

In this work, we proposed a nonlinear optimization-based VIO that can deal with rolling-shutter effects and imperfect camera-IMU synchronization.
We modeled the camera-IMU time offset as a time-varying variable to be estimated.
An efficient algorithm for IMU integration over variable-length time intervals, which is required during the optimization, was also introduced.
The VIO can be robustly launched with our uncertainty-aware initialization scheme.
The experiments demonstrated the effectiveness of the proposed approach.

\bibliographystyle{splncs04}

\end{document}